\documentclass[letterpaper]{article} 
\usepackage{aaai24}  
\usepackage{times}  
\usepackage{helvet}  
\usepackage{courier}  
\usepackage[hyphens]{url}  
\usepackage{graphicx} 
\usepackage{natbib}  
\usepackage{caption} 
\usepackage{algorithm}
\usepackage{algorithmic}

\usepackage{newfloat}
\usepackage{listings}
\DeclareCaptionStyle{ruled}{labelfont=normalfont,labelsep=colon,strut=off} 
\floatstyle{ruled}
\newfloat{listing}{tb}{lst}{}
\floatname{listing}{Listing}
\usepackage{cite}
\usepackage{amsmath,amssymb,amsfonts}
\usepackage{algorithmic}
\usepackage{graphicx}
\usepackage{textcomp}
\usepackage{xcolor}
\usepackage{comment}
\usepackage{multirow}
\usepackage{booktabs}

\AtBeginDocument{
  \setlength\abovedisplayskip{0pt}
  \setlength\belowdisplayskip{0pt}}
\usepackage[margin=0pt, font=footnotesize,labelfont={bf,footnotesize}]{caption}
\usepackage[activate={true,nocompatibility},final,tracking=false,kerning=true,spacing=true,factor=1100,stretch=5,shrink=30]{microtype}

\title{Novel Categories Discovery Via Constraints on Empirical Prediction Statistics}
\author{\begin{tabular}{c}Zahid Hasan\textsuperscript{*}, Abu Zaher Md Faridee\textsuperscript{*}, Masud Ahmed\textsuperscript{*}, Sanjay Purushotham\textsuperscript{*} \\Heesung Kwon\textsuperscript{\$}, Hyungtae Lee\textsuperscript{\$}, Nirmalya Roy\textsuperscript{*}\end{tabular}}
\affiliations{
    \textsuperscript{*}Department of Information Systems, University of Maryland, Baltimore County, USA\\
\textsuperscript{\$}DEVCOM Army Research Laboratory (ARL), USA
}

\usepackage{bibentry}

\begin{document}

\maketitle

\begin{abstract}
\textit{Novel Categories Discovery} (NCD) aims to cluster novel data based on the class semantics of known classes using the open-world partial class space annotated dataset. 
As an alternative to the traditional pseudo-labeling-based approaches, we leverage the connection between the data sampling and the provided multinoulli (categorical) distribution of novel classes. We introduce constraints on individual and collective statistics of predicted novel class probabilities to implicitly achieve semantic-based clustering.
More specifically, we align the class neuron activation distributions under Monte-Carlo sampling of novel classes in large batches by matching their empirical first-order (mean) and second-order (covariance) statistics with the multinoulli distribution of the labels while applying instance information constraints and prediction consistency under label-preserving augmentations. 
We then explore a directional statistics-based probability formation that learns the mixture of Von Mises-Fisher distribution of class labels in a unit hypersphere. 
We demonstrate the discriminative ability of our approach to realize semantic clustering of novel samples in image, video, and time-series modalities. 
We perform extensive ablation studies regarding data, networks, and framework components to provide better insights. 
Our approach maintains $94\%$, $93\%$, $85\%$, and $93\%$ (approx.) classification accuracy in labeled data while achieving $90\%$, $84\%$, $72\%$ and $75\%$ (approx.) clustering accuracy for novel categories in Cifar10, UCF101, MPSC-ARL, and SHAR datasets that match state-of-the-art approaches without any external clustering. 









\end{abstract}

\section{Introduction}

The \textit{Novel Categories Discovery} (NCD) enables developing machine-learning models using partially annotated datasets in an open-world setting~\cite{han2019learning, han2021autonovel, han2020automatically}.
NCD, a specialized case of \textit{Generalized Categories Discovery} (GCD), relaxes the requirement of distinct labels for all underlying classes by assuming a disjoint set between labeled and unlabeled (novel) classes and availability of prior knowledge regarding the number and distribution of novel classes. 
NCD aims to develop a data-driven model that comprehends the general class semantics to simultaneously classify labeled and cluster unlabeled data.

NCD methods employ class notions and semantics derived from labeled (known) classes to cluster unlabeled data.
Prior NCD literature has demonstrated the clustering of novel data via pairwise similarity losses \cite{hsu2017learning}, pseudo-labeling \cite{jia2021joint}, prototype learning \cite{zhang2022automatically}, meta-learning \cite{chi2021meta}, contrastive learning \cite{pu2023dynamic}, kernel network \cite{wang2020open}, Sinkhorn-Knopp clustering \cite{fini2021unified}, variance regularization \cite{hasan2023nev} and self-training \cite{asano2019self} relying novel class uniformity assumption.  
In contrast, we explore the adaptation of the non-contrastive self-supervision within the NCD paradigm to cluster novel data into a fixed number of groups, guided by a prior distribution based on class semantics.


We leverage the probabilistic nature of novel samples and their connection with the multinoulli (categorical) distribution. 
Unlike labeled data with a deterministic target, unlabeled instances are modeled by a multinoulli distribution with known probabilities for the novel classes. 
We exploit this prior in developing a deep-learning-based NCD method that aligns the empirical distribution of the probability matrix obtained from a large batch of Monte Carlo sampled unlabeled data with the given class probabilities to achieve semantic-based clustering of the novel data while simultaneously learning semantics from labeled data.


We rethink NCD from a distribution learning perspective and propose to apply statistical constraints on the predicted probability
to implicitly cluster novel data. 
Our approach shares similarities with canonical correlation analysis \cite{balestriero2023cookbook} and non-contrastive self-supervised methods \cite{balestriero2022contrastive}, as we impose constraints on class-representative neuron activation patterns, aiming to cluster novel class groups separately. Our contributions can be summarized as follows:



\textbf{(1)} 
We propose a novel distribution learning-based approach for NCD that imposes constraints on individual (entropy and consistency) and collective feature statistics (first order, \textit{mean} and second order, \textit{covariance}) of predicted probability. Our instance-wise losses ensure consistent, augmentation-invariant decisions, while feature-wise loss aligns neuron activation patterns with the underlying class distribution.

\textbf{(2)} 
We explore parametric (classifier) and non-parametric class probability estimation. 
The non-parametric method utilizes the prototypes in the unit hypersphere and their angular separation to encode classes. Our prototypical NCD optimizes class-conditioned von Mises Fisher (vMF) distributions for labeled and a mixture of vMFs with the multinoulli probability as mixing parameters for novel classes.


\textbf{(3)} 
We experiment with diverse data modalities, including images, videos, and time series sensor data, to assess our approach. Our in-depth ablation studies cover (i) the dataset; (ii) the architecture; (iii) the framework perspective to report valuable insights into the approach. For reproducibility and engagement, we will open-source our code and data artifacts.



\section{Related works}


Typically NCD approaches follow either a two-stage or a single-stage joint optimization \cite{joseph2022spacing} to facilitate knowledge transfer from labeled to novel data. 
The two-stage approaches can be grouped into learned-similarity and latent-space-based methods \cite{troisemaine2023novel}. The similarity-based NCDs such as CCN \cite{hsu2017learning} and MCL \cite{hsu2019multi} seek to develop a binary classifier that identifies similarity/dissimilarity on unique pairs of instances.  
The latent-space-based approaches such as DTC \cite{han2019learning}, DEC \cite{xie2016unsupervised}, and MM \cite{chi2021meta} aim to learn the latent representation of the high label features from the labeled data and apply a clustering algorithm to cluster novel data.

On the other hand, single-stage methods such as AutoNovel~\cite{han2021autonovel}, UNO~\cite{fini2021unified}, OpenMix~\cite{zhong2021openmix}, \cite{zhang2022automatically}, NEV-NCD \cite{hasan2023nev}  enable NCD to utilize both the labeled and unlabeled data simultaneously with/without self-supervised pretraining. They jointly minimize a combination of supervised (labeled data) and unsupervised losses (unlabeled data). The unsupervised losses are the key to novel data clustering and often rely on clustering-based pseudo-labeling \cite{arazo2020pseudo} and data equipartitioning \cite{fini2021unified}. 

Our single-stage approaches are closely related to and improve generalization upon NEV-NCD \cite{hasan2023nev} addressing the non-uniform class distributions and negative learning redundancy. Moreover, we take inspiration from canonical relationship analysis \cite{hotelling1992relations} based on negative sample-free self-supervised learning \cite{balestriero2023cookbook} such as DINO \cite{caron2021emerging}, Barlow-twin \cite{zbontar2021barlow}, W-MSE \cite{ermolov2021whitening}, SwAV \cite{caron2020unsupervised}.



\section{Learning from Probability Matrix}

\subsection{Notation and problem formulation}


The dataset, $\mathcal{D} = \mathcal{D}^l \cup \mathcal{D}^u$, consists of labeled data $\mathcal{D}^l = \{(\mathbf{x}_i^l, y_{i})\}_{i = 1}^{N^l}$ where $y_{i} \in C_L=\{c_1, ..., c_L\}$ from $L = |C_L|$ classes and unlabeled data $\mathcal{D}^u = \{(\mathbf{x}_i^u)\}_{i = N^l+1}^{N^l + N^u}$ from $U = |C_U|$ non-overlapping latent novel classes $y_i \in C_U=\{c_{L+1}, ..., c_{L+U}\} \wedge C_L \cap C_U = \emptyset$. The novel instances are from a known Multinoulli distribution parameterized by $\mathbf{p}_U = [p_{u, L+1}, ..., p_{u, L+U}]^T$, where the elements $p_{u,j}$ denote the probability corresponding novel classes in $\mathcal{D}^u$. NCD aims to develop models to classify the $\mathcal{D}^l$ by the class semantics and cluster the $\mathcal{D}^u$ based on their latent semantics.

We represent $k$-dimensional vector with all $0/1$ elements as $\mathbf{0}_k/\mathbf{1}_k$.
We define $\mathbf{y}$ as $K = L+U$ dimensional Multinoulli variable with $\{y_j\in\{0,1\}^{L+U} \wedge \sum_{j = 1}^{L+U}y_{j} = 1\}$. The target for the corresponding $i$-th sample is represented by one-hot encoding Multinoulli vector $\mathbf{y}_{i}$ with $j=y_i$-th position $y_{ji} = 1$.
We denote parameterized network encoder $f_\theta$, embedding layer $h_\theta$, and classification head $g_\theta$. The $g_\theta$ consists of $K = L+U$ neurons ($L$ known and $U$ novel) with softmax activation to assign probabilities to each class. 
The predicted class probability for a single instance is denoted by $K \times 1$-dimensional vector $\hat{\mathbf{y}} = g_\theta(h_\theta(f_\theta(\mathbf{x}_i)))$. We also define the probability matrix $\mathbf{P}_{\mathbf{p}} = [\mathbf{p}_1, \mathbf{p}_2, ..., \mathbf{p}_B] \in \mathbb{R}^{K \times B}$ for a batch of $B$ instances by placing $B$ predictions $\{\mathbf{p}_i \subset \mathbb{R}^K, i = 1, 2, ..., B\}$ as column, where $i$-th column, $\mathbf{P}_{\mathbf{p}, *i}$, represents the class probability for a $i$-th instance of $B$.


\subsection{NCD using Multinoulli distribution}
The parameterized discrete $K$-dimensional Multinoulli distribution \cite{murphy2012machine} is the most general distribution over a $K$-way event that describes the possible outcome of one of $K$ mutually exclusive categories with a prior probability. 


\textbf{Definition}: Let $\mathbf{y}= [y_1, ..., y_K]^T$ be a discrete random vector with support $R_\mathbf{y}=\{y_j\in\{0,1\}^K : \sum_{j = 1}^{K}y_{j} = 1\}$. For $K>0$ categories with strictly positive event probabilities $p_1, p_2, ..., p_K$ with $\sum_{i=1}^K p_i = 1$, we define $\mathbf{y}$ with a Multinoulli distribution with parameter $\mathbf{p} = [p_1, ..., p_K]^T \in \mathbb{R}^K$ if its joint probability mass function is: 

\begin{equation}
   Mu(\mathbf{Y}|1, \mathbf{p})= p_{\mathbf{y}}({\mathbf{y}})= 
    \begin{cases}
         \Pi_{j=1}^K p_j^{y_j} & \mathtt{if } \mathbf{y} \in R_{\mathbf{y}}\\
          0 & \mathtt{otherwise}
    \end{cases}
    \label{eq:multi_def}
\end{equation}

The expected value $\mathbb{E}[\mathbf{y}]$ is $\mathbf{p}$ and the covariance matrix $\mathtt{Var}(\mathbf{y})= \Sigma: \Sigma \in \mathbb{R}^{K \times K}$ contains the following elements.
\begin{equation}
    \Sigma_{i,j}= 
    \begin{cases}
          p_i(1-p_i) & \mathtt{if }j=i\\
          -p_ip_j & \mathtt{if } j\neq i
    \end{cases}
\end{equation}


We sample a batch of $B$ labeled/novel instances from  $\mathcal{D}^{l/u}$ using Monte Carlo sampling methods and column-wise stack their corresponding predicted probability over $K=L+U$ classes $\{\mathbf{P}_{\hat{\mathbf{y}},*i}^{l/u} = \hat{\mathbf{y}}_i: i = 1,..., B\}$ for $i$-th instance to form the probability matrix $\mathbf{P}_{\hat{\mathbf{y}}}^{l/u} \subset \mathbb{R}^{K\times B}$.
We will use $\mathbf{P}^{l/u}$ instead of $\mathbf{P}_{\hat{\mathbf{y}}}^{l/u}$ to avoid notation overload in our subsequent discussion.
Assuming a unique class per instance, the discriminative model aims to predict the correct class for the instances by concentrating probability on the representative class neuron. For $\mathbf{P}^l$, we have an exact ground truth matrix $\mathbf{P}_{\mathbf{y}}^l$ that provides a one-hot Multinoulli target variable $\mathbf{y}_i$ for each column ($\mathbf{P}_{*i}^l$) and apply instance-wise supervised objective. 
However, the probability matrix $\mathbf{P}^u$ for novel data samples from $\mathcal{D}^u$ does not have exact instance-wise target $\mathbf{y}_i$ for the predicted columns $\mathbf{P}_{*i}^u$.

Traditional approaches utilize pseudo-labeling to generate Multinoulli targets for the instances and re-train the network to align $\mathbf{P}_{*i}^u$ with the pseudo-target.
Alternatively, we leverage the associated uncertainty and their expected statistical behavior under Monte Carlo sampling by modeling the target $\mathbf{y}$ of each column $\mathbf{P}_{*i}^u$ as a probabilistic sample from a Multinoulli distribution parameterized by $\mathbf{p}_U$. 
The novel sample belongs to one of the novel classes with a Multinoulli probability $\mathbf{p}_U$ and the empirical label distribution from $\mathcal{D}^u$ would follow the distribution. 
Similarly, the distribution of ideal NCD model prediction $\mathbf{P}_{*i}^u$ over the novel data would also align with the same Multinoulli distribution. Based on this phenomenon, we enforce learning to match the novel class Multinoulli distribution defined by $\mathbf{p}_U$ to implicitly cluster the $\mathcal{D}^u$ based on the semantics of $\mathcal{D}^l$. 

To achieve such objectives, we propose statistical constraints (enforce collective feature behavior) and individual constraints (instance-wise) on the $\mathbf{P}^u_{*i}$. 
As statistical constraints, we minimize the distance (divergence) between the empirical distribution of the model prediction for instances from $\mathcal{D}^u$ and the novel class prior distribution. In terms of $\mathbf{P}^u$, we align the empirical mean, $\mathbb{E}[\mathbf{P}^u_{*i}]$, and covariance, $\Sigma^{\mathbf{P}^u} = Var(\mathbf{P}^u) = \mathbb{E}[(\mathbf{P}^u_{*i} - \mathbb{E}[\mathbf{P}^u_{*i}])(\mathbf{P}^u_{*i} - \mathbb{E}[\mathbf{P}^u_{*i}])^T]$ with $\Sigma^{\mathbf{P}^u} \in \mathbb{R}^{K \times K}$, of the model prediction ($\mathbf{P}^u_{*i}$) with the Multinoulli distribution defined by $\mathbf{p}_U$ as follow. 

\begin{equation}
    \mathbb{E}[\mathbf{P}_{*i}^u] = 
    \begin{bmatrix}
           \mathbf{0}_L \\
           \mathbf{p}_U
     \end{bmatrix}
     \label{eq:true_mean}
\end{equation}

\begin{equation}
    \Sigma_{i,j}^{\mathbf{P}^u}= 
    \begin{cases}
            0 & \mathtt{if }i<L+1 \vee j<L+1\\
          p_{i}(1-p_i) & \mathtt{if }j=i \wedge i,j>L\\
          -p_ip_j & \mathtt{if } j\neq i \wedge i,j>L
    \end{cases}
    \label{eq:true_cov}
\end{equation}

We also propose instance-wise regularization to maintain each instance's properties belonging to a single class and decision-invariant under level-preserving augmentations. In terms of $\mathbf{P}^u$, the individual prediction ($\mathbf{P}_{*i}^u$) is expected to maintain a Multinoulli variable property (concentrated probability in the latent class representative neuron as in eq. \ref{eq:multi_def}) and consistent under data augmentations. 

\section{Methodology}

\begin{figure}
 \centering
 \includegraphics[width=1\linewidth]{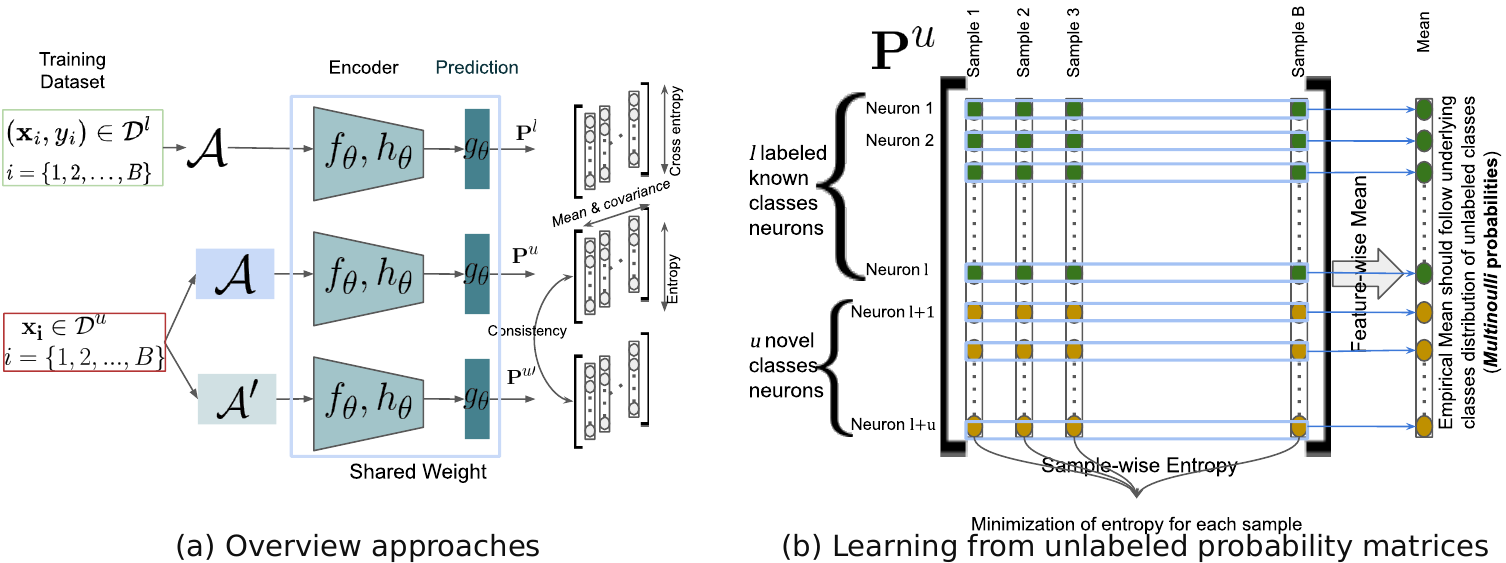}
 \caption{Overview of our NCD approach based on conditions on probability matrix}
 \label{fig:overview_diag}

\end{figure}



\subsubsection{Probability Matrix Formation}


We randomly sample a batch of $B$ instances $\mathbf{x}_i^{l/u}; i =\{1,2...., B\}$ from $\mathcal{D}^{l/u}$ and augment them using random operations from a set of augmentation (label-preserving) operators $\mathcal{A}$. We utilize encoder $f_\theta$ and linear projection $h_\theta$ (without activation) to represent the high dimensional input data in the $d$ dimensional embedding $\mathbf{e}_i^{l/u} =  h_\theta(f_\theta(\mathbf{x}_i^{l/u}): \mathbf{e} \in \mathbb{R}^d$.
Finally, we consider two alternatives of parametric and non-parametric classifiers to form $\mathbf{P}^{l/u}$ from the embedding $\mathbf{e}$.

\textbf{Parametric Classifier with learnable layer:}
The learnable fully connected classification head $g_\theta$ predicts the class distribution for $\mathbf{x}$ over $K\;(=L+U)$ classes $\hat{\mathbf{y}}_i^{l/u}= g_\theta(\mathtt{ReLU}(\mathbf{e}^{l/u})): \hat{\mathbf{y}}_i^{l/u} \subset \mathbb{R}^K$, where, the $L+U$ neurons represent corresponding probability for $K$ classes in a bijection manner. We form probability matrix $\mathbf{P}^{l/u}$ for $B$ labeled/novel samples by $\mathbf{P}_{*i}^{l/u} = \hat{\mathbf{y}}_i^{l/u}; i =\{1,2...., B\}$.

\textbf{Non-parametric Hyperspherical Prototypes:}
In hyperspherical learning, each class is represented by a direction in the unit hypersphere. We randomly select $K$ prototypes $\{\mathbf{\mu}_i | \textit{ } ||\mathbf{\mu}_i|| = 1\}_{i = 1}^K$ in $d$ dimensional hypersphere $\mathbf{\mu}_i \in \mathbb{S}^{d-1} \subset \mathbb{R}^d$ and initialize the prototype matrix $\mathbf{M} \subset \mathbb{R}^{d \times K}: \mathbf{M}_{*i} = \mathbf{\mu}_i$ by column-wise stacking them. 
The higher dimensional ($d>>K$) random sampling ensures their low cosine similarity with high probability \cite{liu2021learning}. 

We find the projected embedding $\mathbf{\Bar{e}}$ in $\mathbb{S}^{d-1}$ by L2 normalization where $\mathbf{\Bar{e}} =  \frac{\mathbf{e}}{||\mathbf{e}||}$. 
The $i$-th instance class assignment depends on the cosine (angular) similarity between its predicted embedding $\mathbf{\Bar{e}}_i$ and the class prototypes. We calculate cosine similarities and apply $\mathtt{Softmax}$ with temperature ($\tau$) scaling to represent the class probability as in eq. \ref{eq:embed_to_p}. Finally, we form $\mathbf{P}^{l/u}$ using the previous strategy with $\mathbf{P}_{*i}^{l/u} = \hat{\mathbf{y}}_i^{l/u}$.

\begin{align} \label{eq:embed_to_p}
    \mathbf{\hat{y}}_i^{l/u} = \mathtt{Softmax}(\frac{1}{\tau}\mathbf{M}^T\mathbf{e}^{l/u}) \\
    \hat{y}_{ik} = \frac{\exp({\frac{1}{\tau}\mathbf{\mu}_k^T \mathbf{\Bar{e}}_i})}{\sum_{k = 1}^K\exp({\frac{1}{\tau} \mathbf{\mu}_k^T \mathbf{\Bar{e}_i}})}
\end{align}

$\mathbf{\hat{y}}_i$ denotes the probability with elements $\hat{y}_{ik}$ that quantify the probability of $i$-th instance to class $k : k \in \{1, ..., K\}$.

\subsubsection{Learning from $\mathbf{P}^l$}
The $\mathcal{D}^l$ provides corresponding exact ground truth $y_i \in C_L$ for instance $\mathbf{x}_i\in \mathcal{D}^u$. 
We utilize the $y_i$ to create a one-hot encoding (Multinoulli variable) vector $\mathbf{y}_i \in \mathbb{R}^{K}$ with element $y_{i, j} =  \mathrm{1}_{[j = y_i]}, j =\{1, ..., K\}, y_j \in C_L$, where, $\mathrm{1}_{[j = y_i]}$ refers identity operators. For $\mathbf{P}^l$, we construct the target matrix $\mathbf{P}_{\mathbf{y}}^l$ with $\mathbf{P}_{\mathbf{y}, *i}^{l} = \mathbf{y}_i$ i.e. the $i$-th column contains the one-hot target for the $i$-th sample of $\mathbf{P}^l$. We minimize instance-wise supervised categorical cross-entropy loss as defined in eq. \ref{eq:ce_sup} to align  $\mathbf{P}^l$ with corresponding $\mathbf{P}^l_{\mathbf{y}}$ by learning the class semantics from $\mathcal{D}^l$.

\begin{align}
 \mathcal{L}_{ce} =  - \underset{(\mathbf{x}_i, \mathbf{y}_i) \in \mathcal{D}^l}{\mathbb{E}} [\frac{1}{B}\sum_{i = 1}^B{\mathbf{P}_{\mathbf{y},*i}^l}^T \ln{\mathbf{P}_{*i}^l} ] \\
 = - \underset{(\mathbf{x}_i, \mathbf{y}_i) \in \mathcal{D}^l}{\mathbb{E}} [\frac{1}{B}\mathtt{tr} ({\mathbf{P}_{\mathbf{y}}^l}^T \ln \mathbf{P}^l)]
 \label{eq:ce_sup}
\end{align}

\textbf{Corollary 1}: In a hyperspherical setting, the $\mathcal{L}_{ce}$ minimization for $\mathcal{D}^l$ optimizes the model to class-conditioned vMF distribution with a selected mean direction $\mu_k: k \in C_L$ and concentration $\kappa = 1/\tau$ (proved in  supplemental section).

\subsubsection{Learning from $\mathbf{P}^u$}

The precise target for the random samples $\mathbf{x}^u \in \mathcal{D}^u$ is unavailable; instead, they are stochastic variables that follow a Multinoulli distribution of $\mathbf{p}_{U}$ over the novel classes.
In particular, the probabilistic target $\mathbf{y}_i|\mathbf{x}^u_i$ for prediction $\mathbf{P}^u_{*i} $ is a sample from $K$-dimensional Multinoulli parameterized by $\mathbf{y}^U = [\mathbf{0}_L^T \; \mathbf{p}_U^T]^T$ over $C_L$ and $C_U$ ($\mathbf{0}_L^T$ represents $0$ probability over the $L$ labeled classes due to disjoint condition) i.e., $\mathbf{y}|\mathbf{x}^u \sim Mu(\mathbf{Y}|1, [\mathbf{0}_L^T \; \mathbf{p}_U^T]^T)$.
Exploiting this prior, we train the parametric model to learn the underlying Multinoulli distribution over novel samples to implicitly cluster $\mathcal{D}^u$ separately based on the class semantics of $\mathcal{D}^l$ by applying instance and feature-wise constraints. 

\textbf{Instance-wise Constraints:}
As instance-wise constraints, firstly, we enforce the Multinoulli variable property of Dirac-delta distribution for each predicted $\mathbf{P}_{*i}^u$. We propose two alternatives to apply such constraint by minimizing either the entropy \cite{grandvalet2004semi} of eq. \ref{eq:ent_reg} or sharpened target self-training \cite{berthelot2019mixmatch} of eq. \ref{eq:shr_reg} with sharpened prediction matrix $\mathbf{P}_{sh} = \mathtt{SoftMax}(\mathbf{P}/s): s <<1 $.

\begin{equation} 
  \mathcal{L}_{H} =  - \underset{\textbf{x}_i \in \mathcal{D}^u}{\mathbb{E}}[\frac{1}{B} \mathtt{tr} ({\mathbf{P}^u}^T\ln{\mathbf{P}^u})]
  \label{eq:ent_reg}
\end{equation}

\begin{equation} 
  \mathcal{L}_{sh} =  - \underset{\textbf{x}_i \in \mathcal{D}^u}{\mathbb{E}}[\frac{1}{B} \mathtt{tr} ({\mathbf{P}_{sh}^u}^T\ln{\mathbf{P}^u})]
  \label{eq:shr_reg}
\end{equation}


Secondly, we enforce augmentation invariant prediction consistency \cite{yang2022divide}. 
We form two prediction matrices $\mathbf{P}^u$ and $\mathbf{P}^{u\prime}$ where their columns $\mathbf{P}_{*i}^u$ and ${\mathbf{P}^{u}_{*i}}^\prime$ represent the prediction for two differently augmented versions of the same $\mathbf{x}_i$. 
We consider two alternatives as consistency loss by either minimizing the Frobenius norm (defined by $||\mathbf{A}||_F = \sqrt {\mathtt{tr} (\mathbf{A}^T\mathbf{A})}$) of the difference matrix between $\mathbf{P}^u$ of eq. \ref{eq:mse} and $\mathbf{P}^{u\prime}$ or swapped prediction where prediction of $\mathbf{P}^u$ act as a target for $\mathbf{P}^{u\prime}$ and vice-versa of eq. \ref{eq:swapped}.


\begin{equation}
 \mathcal{L}_{mse} = \underset{\mathbf{x}_i \in \mathcal{D}^{u/l}}{\mathbb{E}} [\frac{1}{B}||\mathbf{P}^{u}  - {\mathbf{P}^{u}}^\prime ||_F] 
 \label{eq:mse}
\end{equation}

\begin{equation}
 \mathcal{L}_{sp} = - \frac{1}{B} \underset{\mathcal{D}^{u}}{\mathbb{E}}  [{\mathtt{tr}(\mathbf{P}^{u}}^T  \ln{\mathbf{P}^{u}}^\prime) + \mathtt{tr}({{\mathbf{P}^{u}}^\prime}^T  \ln {\mathbf{P}^{u}})] 
 \label{eq:swapped}
\end{equation}

\textbf{Feature-wise Constraints}
They enforce the statistical (collective) behaviors of $\mathbf{P}^u$ columns to follow the prior Multinoulli distribution with $\mathbf{y}^U = [\mathbf{0}_L^T \; \mathbf{p}_U^T]^T$ by aligning the empirical mean and covariance of the $\mathbf{P}^u_{*i}$ with the objective distribution. 
As the sample mean of a Multinoulli represents the empirical probability, we minimize their KL-divergence as in eq. \ref{eq:f-div-loss-mean} to align the true distribution defined in eq. \ref{eq:true_mean} and empirical mean calculated using eq. \ref{eq:emp_mean}. 
Besides relying on instance-wise loss to emerge cluster by avoiding the trivial solution of fixed mean values, we also regularize the estimated empirical covariance matrix \ref{eq:cov-matrix} by minimizing the Frobenius norm of difference matrix between empirical and true covariance of eq. \ref{eq:true_cov} as in eq. \ref{eq:cov-loss} \cite{hasan2023nev}.

\begin{equation}
    \hat{\mathbb{E}}[\hat{\mathbf{y}}] = \hat{\mathbb{E}}[\mathbf{P}^u_{*i}]=   \frac{1}{B}\sum_{i = 1}^B\mathbf{P}_{*i}^u= \frac{1}{B}\mathbf{P}^u\mathbf{1}_B
    \label{eq:emp_mean}
\end{equation}

\begin{equation}
 \mathcal{L}_{kl-div} =  \underset{\mathbf{x}_i \in \mathcal{D}^u}{\mathbb{E}} {\mathbf{y}^U}^T[\ln{\mathbf{y}^U} - \ln{(\frac{1}{B}\mathbf{P}^u\mathbf{1}_B)]}
 \label{eq:f-div-loss-mean}
\end{equation}

\begin{align}
    \mathtt{Zero~centered}~\mathbf{P}^u =  \{\mathbf{\Bar{P}}^u: {\mathbf{\Bar{P}}^u_{*i}} = {\mathbf{P}^u_{*i}  - \hat{\mathbb{E}}[\hat{\mathbf{y}}]}\} \\
    \hat{\Sigma}^{\mathbf{P}^u} = \frac{1}{B} \mathbf{\Bar{P}}^{u}\Bar{\mathbf{P}}^{uT} \;\mathtt{where~} \hat{\Sigma}^{\mathbf{P}^u} \in \mathbb{R}^{K \times K}
    \label{eq:cov-matrix}
\end{align}

\begin{equation}
    \mathcal{L}_{var} = || \hat{\Sigma}^{\mathbf{P}^u} - \Sigma^{\mathbf{P}^u}||_F
    \label{eq:cov-loss}
\end{equation}

\textbf{Optimization:}
We jointly optimize the weighted ($\lambda$) sum of the discussed losses \ref{eq:joint_op1} alternatively utilizing $\mathbf{P}^l$ and $\mathbf{P}^u$. 


\begin{equation}
 \mathcal{L}= 
 \begin{cases}
 \lambda_{ce}\mathcal{L}_{ce} & \mathtt{if } \mathbf{x} \in \mathcal{D}^l\\ \lambda_u\mathcal{L}_{u} & \mathtt{if } \mathbf{x} \in \mathcal{D}^u
 \end{cases}
 \label{eq:joint_op1}
\end{equation}
\begin{equation}
 \mathcal{L}_{u}= \lambda_{H}\mathcal{L}_{H/sh} + \lambda_{m}\mathcal{L}_{mse/sp} + 
 \lambda_{kl}\mathcal{L}_{kl-div} + \lambda_{v}\mathcal{L}_{var}
 \label{eq:joint_op}
\end{equation}

\textbf{Corollary 2}: In the hyperspherical setting, the joint optimization of $\mathcal{L}_{H}$ and $\mathcal{L}_{kl-div}$ over $\mathcal{D}^u$ optimize the model to fit the class embedding in a mixture of $U$ vMFs with unique mean directions parameterized by $\{(p_{L+1}, \mu_{L+1}, \kappa), ..., (p_{L+U}, \mu_{L+U}, \kappa)\}$ with the mixing parameters follows the novel class multinoulli distribution (proved in  supplemental section).

\section{Experiments}

We demonstrate the applicability of our NCD framework by experimenting with image, video, and time-series data modalities and performing extensive ablation studies.

\subsection{Dataset}
\paragraph{}
\textbf{CIFAR10}~\cite{krizhevsky2009learning} dataset contains $32 \times 32 \times 3$ images of $10$ classes with  $5000$ training and $1000$ testing images per class. Besides the ablation studies, we follow the setting of \cite{zhang2022automatically} to benchmark.

\textbf{UCF101} We experiment with the UCF101~\cite{soomro2012ucf101} dataset with randomly selected $90$ labeled and the remaining $11$ as novel actions.

\textbf{MPSC ARL dataset:} ~\cite{hasan2023nev} It consists of three synchronous views from three realistic camera positions with varying camera settings. It contains class-balanced ten regular actions with static (sitting, standing, lying with face up and down) and dynamic patterns (walking, push up, waving hand, leg exercise, object carrying, and pick/drop).

\textbf{UniMiB SHAR:}~\cite{micucci2017unimib} dataset has 17 physical activities grouped into (1) nine types of activities of daily living (ADL) and (2) eight types of falls from 30 subjects with acceleration sensor being placed in left and right trouser pockets resulting $11,771$  samples. Here, we only focus on the nine ADL types.

\begin{table}[!htb]
\caption{Labeled/Unlabeled splits for NCD experiments}
\label{table:data_split}
\centering
\footnotesize
\resizebox{.9\columnwidth}{!}{
\begin{tabular}{@{}ccccc@{}}
\toprule
Dataset & Cifar10   & UCF   & MPSC-ARL    & SHAR  \\ \midrule
\begin{tabular}[c]{@{}c@{}}Data Split\\ (labeled/unlabeled)\end{tabular} & \begin{tabular}[c]{@{}c@{}}4/6, 5/5\\ 6/4\end{tabular} & 90/11 & \begin{tabular}[c]{@{}c@{}}4/6, 5/5\\ 6/4\end{tabular} & \begin{tabular}[c]{@{}c@{}}5/4\\ 6/3\end{tabular} \\ \bottomrule
\end{tabular}
}
\end{table}



\subsection{Implementation Details}

\textbf{Network Architectures:} For image dataset we experimented with ResNet (ResNet18, ResNet34, ResNet50)~\cite{he2016deep} and MobilenetV2~\cite{sandler2018mobilenetv2} architectures. SlowFast~\cite{feichtenhofer2019slowfast}, R2+1D ~\cite{tran2015learning}, and P3D~\cite{qiu2017learning} are used as the video action recognition backbone. We re-purpose the recently proposed CoDEm~\cite{faridee2022codem} pipeline to experiment with the SHAR dataset.

\textbf{Optimization:}
We apply fixed and adaptive weights \cite{liu2022residual, hasan2023nev} for loss components and optimize the joint loss objectives using the stochastic gradient descent (SGD) optimizer with a linear learning rate scheduler. 
In our experiments, we fix $\lambda_{H} = 1.0$ and other hyper-parameters ($\tau = 0.05$, $Sr = 0.1$). 
To calculate the empirical mean and covariance, we trade-off between computational cost and the law of large numbers maintaining a batch size of a minimum of $10 \times U$ (batch of $256$, $128$, $64$, and $256$ for Cifar10, UCF101, MPSC-ARL, and SHAR).
We randomly sample our initial prototype centers and fix the concentration parameter to $10 (\tau = 0.1)$. 
In our setting, the concatenation of $\mathbf{P}^u$ and $\mathbf{P}^{u\prime}$ to calculate the class activation order statistics performed better.
Moreover, we consider the data imbalance to prevent bias and overfit the model towards the majority sampled group due to equal sampling from $\mathcal{D}^l$ and $\mathcal{D}^u$ at each iteration (e.g., $90/11$ labeled/unlabeled separation in the UCF101 created an imbalance) and tackle it by making proportionate batch sizes. 
During the joint optimization, we observe the behavior of individual loss components to ensure their similar scale and simultaneous convergence without adversarial impact on each other.



\subsection{Baseline Comparison and Ablation Studies}

Comparing NCD methods is challenging due to inherent variations in the backbone architectures, usage of datasets, self-supervised pre-training, label-unlabeled class selection, assumptions of data, applications of data augmentations, transfer learning, and training complexity. 
In our evaluation, we primarily focus on pre-training free methods using convolutional architectures and showcase the effectiveness of our simplistic NCD scheme across multiple domains.
We compare the performance of our proposed method with several baselines: (\textbf{i)} naive supervised baseline of $\mathcal{D}^l$, \textbf{(ii)} the state-of-the-art Sinkhorn-Knopp clustering \cite{asano2019self} with pseudo-labeling: UNO~\cite{fini2021unified}, and two earlier two-stage benchmark methods: \textbf{(iii)} KCL \cite{hsu2017learning} and \textbf{(iv)} MCL \cite{hsu2019multi}, using the identical Cifar10 setting.


We ablate over different design choices of NCD settings by utilizing data domain knowledge to better understand the performance characteristics of our model.
Our ablation experiments can be broadly categorized into three main groups:

\textbf{(1) } From a \textbf{dataset} perspective, we perform three sets of experiments: (i) Augmentation $\mathcal{A}$: We vary the numbers and strength of $\mathcal{A}$. (ii) Label/Unlabeled classes $C_L/C_U$: 
We consider different numbers and sets of $C_L$ and $C_U$ by selecting the class groups with varying semantic notions randomly and manually to observe the impacts of labeled sets. 
In the manual settings, we considered animal classes as labeled data (bird, cat, deer, dog, frog, horse) and attempted to discover machine classes \textit{plane, car, ship, truck}) and considered static classes as $C_L$, dynamic classes as $C_U$, and vice-versa for Cifar10 and MPSC-ARL datasets respectively.
(iii) Distribution $\mathbf{p}_U$: We vary the $\mathbf{p}_U$ beyond the uniform setting. 

\textbf{(2)} From a \textbf{network architecture} perspective, we analyze the NCD performances for the different convolutional architecture-based $f_\theta$ with varying strengths, as discussed above.

\textbf{(3)} Finally, from the proposed \textbf{framework} perspective, we investigate the importance of individual loss components and the framework's robustness by varying the hyperparameters, e.g., loss weights, optimization approaches, and batch sizes.








\section{Results}

We evaluate our methods in left-out labeled and novel instances using quantitative and qualitative metrics. We scrutinize the quality of the embeddings utilizing low-dimensional t-SNE visualization.
We quantified the performance over the labeled and unlabeled data by deploying standard accuracy (top-1) and average clustering accuracy (ACC) (eq. \ref{eq:acc_mat}), respectively. We report the average accuracy and ACC over the three runs from different initializations.

\begin{equation}
 ACC = \underset{p \in Perm}{\max} \frac{1}{N} \sum_{i = 1}^N \mathrm{1} \{y_i = p({\hat{y}_i})\}
 \label{eq:acc_mat}
\end{equation}
\noindent Here $ y_i \in C_U $ and $\hat{y}_i$ denote the label and predicted label of $\mathbf{x}_i \in \mathcal{D}^u$. $Perm$ set contains all permutations for labels of $\forall \mathbf{x}_i$ computed efficiently using \cite{kuhn1955hungarian}.

\subsection{Main Results}

We perform quantitative analysis for labeled and novel instances over the left-out data of the image (Cifar10), video (UCF101 and MPSC-ARL), and time-series sensor (SHAR). Our distribution learning approach learns label semantic-based clustering for novel categories of $\mathcal{D}^u$ (high ACC) while classifying the $\mathcal{D}^l$ (high accuracy). We report the performance (both parametric and hyperspherical) for image data in Table~\ref{table:cifar10_main_res} and others in Table~\ref{table:result_ucf101}. In all cases, our method performs on par with state-of-the-art NCD approaches and outperforms the close-set setting in semantic-based novel pattern discovery. Moreover, our hyperspherical prototype outperforms the parametric classifier model consistently. 

\begin{table}[!htb]

\centering
\small
\caption{Comparative performance for classification of labeled and clustering novel categories on CIFAR10 test dataset with our approach (ResNet50) with 5/5 labeled/unlabeled setting.}
\label{table:cifar10_main_res}
\resizebox{.6\columnwidth}{!}{
\begin{tabular}{@{}lcc@{}}

\toprule
Methods   & $\mathcal{D}^l$ (Accu \%)       & $\mathcal{D}^l$  (ACC \%)       \\ \midrule
KCL       & 79.4 $\pm$ 0.6 & 60.1 $\pm$  0.6 \\
MCL       & 81.4 $\pm$ 0.4 & 64.8 $\pm$ 0.4  \\
DTC       & 58.7 $\pm$ 0.6 & 78.6 $\pm$ 0.2  \\
AutoNovel & 90.6 $\pm$ 0.2 & 88.8 $\pm$ 0.2  \\
Ours      & 90.2 $\pm$ 0.5 & 85.5 $\pm$ 0.6  \\ 
Ours (prototype)      & 90.3 $\pm$ 0.5 & 87.2 $\pm$ 0.4  \\\bottomrule
\end{tabular}}
\end{table}

\begin{table}[!htb]
\centering
\small
\caption{Quantative and comparative analysis on UCF101 (90/11) and MPSC-ARL dataset (5/5) with R2+1D networks and SHAR dataset. The rows represent Accuracy and ACC respectively.}
\label{table:result_ucf101}
\resizebox{.7\columnwidth}{!}{
\begin{tabular}{@{}ccccccc@{}}
\toprule
   &    Ours & Proto & NEV & UNO  & SpCn & Sup \\ \midrule
\multirow{2}{*}{UCF}              & 92.3 & 92.4 & 92.4   & 92.1 & 93.2 & 92.6      \\
    & 86.5 & 88.2  & 82.7   & 79.8 & 24.1 & 16.4      \\ \midrule
\multirow{2}{*}{\begin{tabular}[c]{@{}c@{}}MPSC-\\ ARL\end{tabular}} & 82.3 & 82.4 & 82.9    & 83.4 & 85.2 & 84.6      \\
 & 70.2 & 74.5 & 67.3    & 62.1 & 25.8 & 26.3      \\ \midrule
\multirow{2}{*}{\begin{tabular}[c]{@{}c@{}}SHAR \end{tabular}}  & 93.2 & 95.5 & 92.8   & 90.5 & 94.2 & 92.2     \\
  & 70.2 & 75.01 & 68.5   & 64.2 & 35.2 & 30.1      \\ \bottomrule 
\end{tabular}
}
\end{table}

\begin{figure*}[!htb]
    \centering
    \includegraphics[width = .9\linewidth]{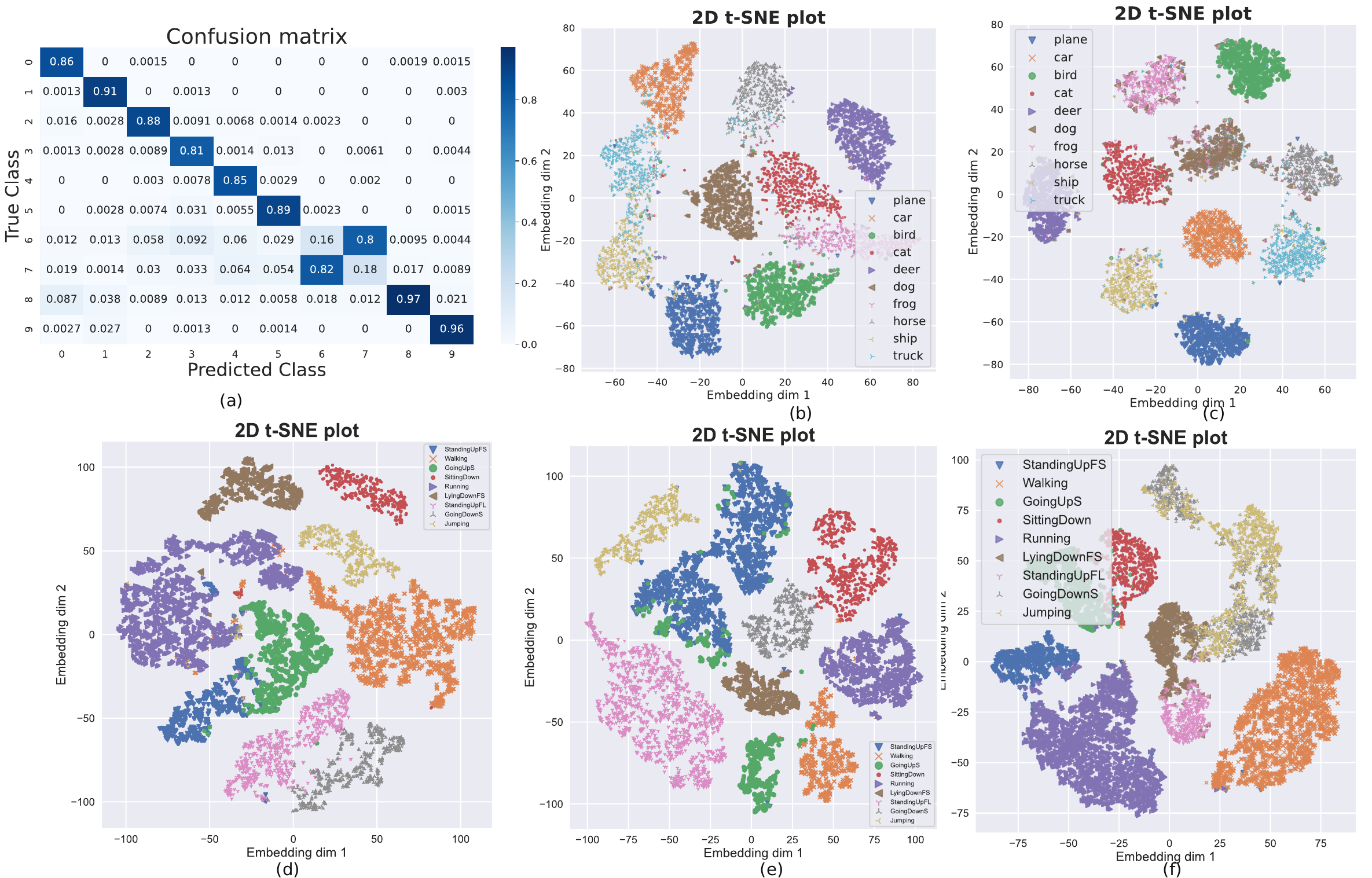}
    \caption{(a) Confusion matrix for Cifar10 dataset. The data embedding separately clusters the label and novel data groups using the ResNet50 model with on 5/5 split cifar10  (b) classifier layer (c) Hyperspherical embedding.  The SHAR dataset embedding separation case for two different L/U combinations (d),(e) and a failed case due to close time-series data pattern (f)}
    \label{fig:ts_cf_cifar}
    \vspace{-0.2in}
\end{figure*}

We also report a sample confusion matrix at Figure~\ref{fig:ts_cf_cifar}(a). it demonstrates that our model assigns one-to-one mapping between class representative neurons to novel classes and avoids class collapse for the $\mathcal{D}^u$. 
We further analyze the embedding and observe our model successfully learns to distinguish and cluster the novel classes from labeled class clusters in the embedding space as shown in Figure~\ref{fig:ts_cf_cifar}(b).  We also report better clustering with higher inter-class separability and intra-class similarity with the hyperspherical prototypes in the time series domain  as shown in Figure~\ref{fig:ts_cf_cifar}(d),~\ref{fig:ts_cf_cifar}(e) and~\ref{fig:ts_cf_cifar}(f).


\subsection{Ablation Study}

\textbf{Data Perspective:}
Data augmentation provides the missing inductive bias to identify class-specific features to make the model invariant to label-preserving noises by providing multiple views \cite{cabannes2023ssl}. 
Learning consistency across multiple augmented versions of the same instance enforces the network to understand the class semantics and prevents overfitting random data properties. 
The NCD model suffers from performance drops in all modalities in the absence of strong augmentation schemes as reported in Table \ref{table:data_ablate_cifar} and \ref{table:data_ablate_mpsc-arl} as they provide extended of missing inductive biases compared to weaker augmentations. 

The labeled class semantics and the relative number of label/unlabeled classes play a vital role in clustering $\mathcal{D}^u$ shown in Table \ref{table:data_ablate_cifar} and \ref{table:data_ablate_mpsc-arl}. 
When the class semantic varies across $\mathcal{D}^l$ and $\mathcal{D}^u$ (animal/machine separation in Cifar10 and static/dynamic separation in the MPSC-ARL) the NCD performance drops as the labeled class fails to provide uniform class notion for novel categories due to semantic shift. We observe similar patterns in SHAR as the model performance degrades as their embedding mixes as shown in figure \ref{fig:ts_cf_cifar} (f). 
Further, NCD clustering performance degrades as the proportional number of novel classes increases. 

Finally, our proposed method performs robust clustering of novel data clustering performance and tackles various class distributions beyond uniform class distribution. However, the model performance degrades under extremely skewed tail class distribution due to overfitting issues as the network biases towards the head novel classes.

\begin{table}[!htb]
\centering

\caption{Ablation study (dataset perspective) on Cifar10 test dataset with our approach ResNet18 setting}
\label{table:data_ablate_cifar}
\resizebox{.9\columnwidth}{!}{%
\begin{tabular}{@{}lccc@{}}
\toprule
           & Description   & $\mathcal{D}^l$ (Accu \%)    & $\mathcal{D}^l$  (ACC \%)   \\ \midrule
\multirow{2}{*}{Augmentation}      & Strong    & 85.2 & 81.82 \\
           & Weak      & 84.6 & 72.1  \\ \midrule
\multirow{2}{*}{\# L/U Classes}    & 6/4       & 85.1 & 84.4  \\
           & 4/6       & 86.5 & 76.2  \\ \midrule
\multirow{2}{*}{\begin{tabular}[c]{@{}l@{}} L/U  classes\end{tabular}} & \begin{tabular}[c]{@{}l@{}}Animal/machine\end{tabular} & 86.2 & 62.5  \\
           & \begin{tabular}[c]{@{}l@{}}Machine/Animal\end{tabular} & 85.7 & 60.3  \\ \midrule
\multirow{3}{*}{Distribution $\mathbf{p}_U$}      & $[1/3, \{1/6\}^4]$        & 85.5 & 81.4  \\
           & $[3/7, \{1/7\}^4] $    & 85.4 & 80.7  \\
           & $[1/2, \{1/8\}^4] $  & 85.7 & 75.1  \\ \bottomrule 
\end{tabular}
}
\end{table}

\begin{table}[!htb]
\centering
\footnotesize
\caption{Ablation study (dataset perspective) on MPSC-ARL test person dataset with our approach R2+1D setting}
\label{table:data_ablate_mpsc-arl}
\resizebox{.9\columnwidth}{!}{%
\begin{tabular}{@{}lccc@{}}
\toprule
          & Description    & $\mathcal{D}^l$ (Accu \%)    & $\mathcal{D}^l$  (ACC \%) \\ \midrule
\multirow{2}{*}{Augmentation}   & Strong         & 82.3 & 70.2 \\
          & Weak           & 80.9 & 60.5 \\ \midrule
\multirow{2}{*}{\# L/U Classes} & 6/4            & 82.1 & 72.3 \\ 
          & 4/6            & 82.5 & 65.2 \\  \midrule
\multirow{2}{*}{L/U Classes}    & Dynamic/Static & 82.1 & 65.1 \\
          & Static/Dynamic & 83.5 & 58.6 \\ \midrule
\multirow{3}{*} {Distribution $\mathbf{p}_U $}   &  $[1/3, \{1/6\}^4]$  & 82.1 & 67.6 \\  
          & $[3/7, \{1/7\}^4] $   & 82.4 & 65.8 \\
          & $[1/2, \{1/8\}^4 ]  $  & 82.4 & 62.4 \\ \bottomrule 
\end{tabular}
}
\end{table}


\textbf{Model Perspective}
We observe that the deeper CNN model improves the classification and clustering accuracy simultaneously in both $\mathcal{D}^l$ and $\mathcal{D}^u$ in Table \ref{table:model_ablate}. 
However, after certain strengths compared to the dataset, CNN-based models' performance reaches a plateau. 

\textbf{Framework Perspective:}
Across modalities, we observe similar impacts of the individual loss objects as reported in Table~\ref{table:framework_ablate_cm}. The NCD performance relies on $\mathcal{L}_{ce}$ to learn the semantic notion of class formation and achieve clustering based on random features with their presence. The $\mathcal{L}_{mse}$ loss ensures consistency by incorporating key-semantic information in clustering. Presence of $\mathcal{L}_{mse}$ prevents the model from clustering the data based on the noise properties. It also helps the model to identify and differentiate key-semantic features across multiple augmented versions.

\begin{table}[!htb]
\centering
\footnotesize
\caption{Ablation study (model variation) for classification of labeled and clustering novel categories on test dataset with our approach with 5/5 labeled/unlabeled setting}
\label{table:model_ablate}
\resizebox{.8\columnwidth}{!}{
\begin{tabular}{@{}llcc@{}}
\toprule
Dataset         & Network     &  $\mathcal{D}^l$ (Accu \%)       & $\mathcal{D}^l$  (ACC \%)        \\ \midrule
\multirow{4}{*}{Cifar10}   & ResNet50    & 90.2 $\pm$ 0.5 & 85.5  $\pm$  0.6 \\
      & ResNet34    & 87.51       & 81.93       \\
      & ResNet18    & 85.20       & 81.82       \\
      & MobileNetV2 & 84.77       & 80.14       \\ \midrule
\multirow{2}{*}{UCF101}         & SlowFast    & 92.8        & 88.7        \\
      & R2 +1D      & 92.2        & 86.5        \\
      \midrule
\multirow{3}{*}{\begin{tabular}[c]{@{}l@{}}MPSC-\\ ARL\end{tabular}} & SlowFast    & 85.2        & 72.3        \\
      & R2+1D       & 82.3        & 70.2        \\
      & P3D         & 74.2        & 61.2        \\ \bottomrule

\end{tabular}
}
\end{table}

Our approach avoids trivial solutions by balancing between two extreme collapses: the model output predicts the $\mathbf{p}_U$ along the novel class representative neurons or predicts a fixed class regardless of the input.
The first-order statistics-based distribution learning loss $\mathcal{L}_{kl}$ prevents one class prediction from dominating but encourages the collapse to the fixed multinoulli distribution probability prediction, meanwhile the information constraint entropy loss  $\mathcal{L}_{H}$ has the opposite effect that collapses all novel classes into a single cluster. 
Finally, the $\mathcal{L}_{kl}$ and $\mathcal{L}_{var}$ complement each other to some extent, and their simultaneous feedback provides the best distribution and cluster learning performance across modalities. 

\begin{table}[!htb]
\centering
\footnotesize
\caption{Ablation study (loss function) on Cifar10 and MPSC-ARL test dataset with our approach (`x' denotes the loss is disabled). }
\label{table:framework_ablate_cm}
\resizebox{.9\columnwidth}{!}{%
\begin{tabular}{@{}cccccccc@{}}
\toprule
$\mathcal{L}_{kl}$ & $\mathcal{L}_{var}$ & $\mathcal{L}_{mse}$ & $\mathcal{L}_{H}$ & $\mathcal{L}_{ce}$ & \begin{tabular}[c]{@{}l@{}}Cifar10\\ ACC \%\end{tabular} & \begin{tabular}[c]{@{}l@{}}MPSC\\ ACC \%\end{tabular} & \begin{tabular}[c]{@{}l@{}}SHAR\\ ACC\%\end{tabular} \\ \midrule
      &        &        &      & x     & 43.5      & 37.8  & 35.5 \\
      &        &        & x    &       & 25.5      & 31.2   & 34.2\\
      &        & x      &      &       & 53.9      & 48.5   & 40.6\\
      & x      &        &      &       & 75.8      & 62.7   & 65.2\\
x     &        &        &      &       & 66.1      & 57.3   & 59.8\\
x     & x      &        &      &       & 26.7      & 29.5   &  31.8 \\ \bottomrule   
\end{tabular}
}
\end{table}

\section{Discussion and future works}

Our approach underperforms when the class semantics change between labeled and novel classes and in the case of extreme tails distribution. 
It also relies heavily on the availability of a number of novel classes and their distribution and degrades with their assumed prior deviates from the true settings. 
We discuss our findings of three such cases for the cifar10 and SHAR datasets (complete experiments and results are in the supplementary section). 

\textbf{Case 1} (Distribution prior mismatch): The performance degrades as the assumed and actual distributions diverge. 
However, we have found uniform prior works as a good choice and the tail classes suffer the most. 

\textbf{Case 2} (Underestimating the class number): Since we already perform minimal clustering \cite{wang2022rethinking}, operating beyond minimal clustering results in under-clustering that causes the class collapse in the representation layers and a significant performance drop.
However, the model tends to form super-clusters by merging the novel classes with close semantics. 

\textbf{Case 3} (Overestimation of class number): As we consider each cluster as individual classes without any connection in between, it enforces the model to over-cluster the data relying on some class irrelevant dominant feature basis and causes severe inter-class separation.

Although it may seem intuitively beneficial to over-cluster to learn more features similar to contrastive self-supervised learning, it degrades the model performance.
The contrastive learning does not enforce strict cluster numbers assuming that the clusters are subsets of the same latent classes \cite{arora2019theoretical}. 
Meanwhile, we enforce strict cluster numbers with some prior relative cluster size. 
In summary, the erroneous class number (major factor) and distribution assumption negatively impacts our novel clusters and labeled classes and their learning behavior depends on the dataset, model, and relative weights of individual losses.
In such cases, the $\mathcal{L}_{kl}$ deems inconsistent with the class notation and acts as the noisy objectives in the joint optimization causing performance degradation \cite{tanaka2018joint}.







In the future, we plan to investigate details about these cases and explore potential solutions to enhance robustness towards GCD \cite{vaze2022generalized} problems by relaxing NCD distribution assumptions and identifying the detailed impacts of individual components, e.g., models, data, modalities, and frameworks in discovering novel categories. 
Further, we plan to enhance semantic-based clustering in the future by tackling the problem of initial cluster corruption during the learning phase by applying label correction mechanisms to fine-tune the model performance. 
Finally, we aim to investigate distribution learning frameworks toward a theoretical understanding and generalized formations by bridging its connection to canonical correlation analysis-based self-supervision as our method seeks to cluster the novel data without using specific negative samples similar to alignment-uniform learning \cite{wang2020understanding} of self-supervision.

\section{Conclusion}

We propose a novel NCD approach that enforces Multinoulli distribution learning for the class activation patterns. We apply joint constraints on first-order and second-statistics of class representative neurons, information of individual instances, and classification loss to learn a single network entailing semantic-based partitions of labeled and novel data. 
We evaluate our approach by performing extensive experiments with partial class-space annotated image, video, and time-series datasets and achieve compatible performance with state-of-the-art NCD works without any additional pertaining and pseudo-labeling
Further, we analyze the importance of class notions, augmentations, sampling, and individual losses toward realizing semantic-based clustering.
\section*{Acknowledgement}
We acknowledge the support of DEVCOM Army Research Laboratory (ARL) and U.S. Army Grant No. \texttt{W911NF21-20076}.

\bibliography{refs}

\end{document}


\maketitle

\section{Theoretical Underpinning for Hyperspherical Embedding}

\textbf{Preliminaries}: In directional statistics,\cite{mardia2000directional} von Mises-Fisher (vMF) distribution
defines a probability distribution over the points in the unit sphere. A $d$-dimensional ($d\geq 2$) normalized random vector $\mathbf{x}$ on the surface of unit sphere $\mathbb{S}^{d-1} = \{\mathbf{x} \subset \mathbb{R}^d: ||\mathbf{x}||_2 = 1\}$ follows a $d$-variate vMF distribution parameterized by mean $\mu: ||\mu||_2 =1$ and concentration $\kappa: \kappa \geq 0$ if its probability density function is given by eq. \ref{eq:vmf}.

\begin{equation} \label{eq:vmf}
  f(\mathbf{x}| \mathbf{\mu}, \kappa) =  C_{d} (\kappa) \exp (\kappa \mathbf{\mu}^T \mathbf{x})
\end{equation}

The $C_d(\kappa)$ denotes normalization factor defined as \ref{eq:vmf_norm}

\begin{equation} \label{eq:vmf_norm}
    C_d(\kappa) =  \frac{\kappa^{0.5d-1}}{(2\pi)^{0.5d} I_{0.5d-1}(\kappa)}
\end{equation}

Where $I_\nu(a)$ denotes a modified Bessel function of the first kind with order $\nu$ and argument $a$. The concentration parameter $\kappa$ is the inverse of variance that determines the concentration of probability mass around the $\mu$.

The mixture of $K$ vMF distributions (movMF) \cite{banerjee2005clustering} with probability distribution given by equation \ref{eq:movmf} models multimodal distribution on the unit-hypersphere. 

\begin{equation}
    f(\mathbf{x}| \Theta) =  \sum_{i = 1}^K \pi_i C_{d} (\kappa_i)~\exp (\kappa_i \mathbf{\mu}_i^T \mathbf{x})
    \label{eq:movmf}
\end{equation}

The parameter $\Theta = \{(\pi_1, \mu_1, \kappa_1), ..., (\pi_K, \mu_K, \kappa_K)\}$ gathers the mixing parameters, mean directions, and concentration parameters.
The $\pi_k$ are called the mixing parameters with $\sum_{i=1}^K \pi_i = 1 : \pi > 0 $ that determine the mixing proportion of $k$-th vMF distribution.

\textbf{Corollary 1}: The $\mathcal{L}_{ce}$ minimization for $\mathcal{D}^l$ optimizes the model to class-conditioned vMF distribution with a selected mean direction $\mu_k: k \in C_L$ and concentration $\kappa = 1/\tau$.

\textbf{Proof}:
For the instances from the labeled set, we utilize their corresponding labels to train the model to minimize the cross-entropy between the ground truth and predicted probabilities over the labeled classes from $C_U$. The instance-wise cross-entropy loss $\mathcal{L}_{ce, i}$ minimization is equivalent to maximizing the alignment between the target class prototype and the instance embedding as shown below. 

\begin{align*}
\begin{split}
    \mathcal{L}_{ce, i } & = - \mathbf{y}_i^T \ln (\mathbf{\hat{y}}_i) \\
    & = - \sum_{l = 1}^K 1_{[y_l = k]} \ln \frac{\exp(\frac{1}{\tau} \mathbf{\mu}_l^T\mathbf{\Bar{e}_i)}}{\sum_{j =1}^U\exp(\frac{1}{\tau} \mathbf{\mu}_j^T\mathbf{\Bar{e}_i)}} \mathtt{ if } \mathbf{x}_i \in c_k  \\
    & = - \ln \frac{\exp(\frac{1}{\tau} \mathbf{\mu}_k^T\mathbf{\Bar{e}_i)}}{\sum_{j =1}^U\exp(\frac{1}{\tau} \mathbf{\mu}_j^T\mathbf{\Bar{e}_i)}}
\end{split}
\end{align*}

The instance loss is minimized with $\frac{\exp(\frac{1}{\tau} \mathbf{\mu}_k^T\mathbf{\Bar{e}_i})}{\sum_{j =1}^U\exp(\frac{1}{\tau} \mathbf{\mu}_j^T\mathbf{\Bar{e}_i)}} \approx 1$ satisfying $\exp(\frac{1}{\tau} \mathbf{\mu}_k^T\mathbf{\Bar{e}_i}) >> \sum_{j, j \neq k}^K\exp(\frac{1}{\tau} \mathbf{\mu}_j^T\mathbf{\Bar{e}_i})$ when the embedding align with the class mean direction, eventually maximizing the class conditioned vMF probability. 
This exact loss formulation based on vMF distribution under equal privilege assumption \cite{hasnat2017mises} has previously demonstrated success in supervised \cite{hasnat2017mises} and out-of-distribution \cite{ming2022exploit} learning.

\textbf{Corollary 2}: The joint optimization of $\mathcal{L}_{H}$ and $\mathcal{L}_{kl-div}$ over $\mathcal{D}^u$ train the model to fit the class embedding in a mixture of $U$ vMFs with unique mean directions parameterized by $\{(p_{L+1}, \mu_{L+1}, \kappa), ..., (p_{L+U}, \mu_{L+U}, \kappa)\}$ with the mixing parameters follows the novel class multinoulli distribution.

\textbf{Proof}:
Entropy loss $\mathcal{L}_H$ is minimized when the discrete probability distribution takes the form of a multinoulli variable with concentrated probability at one point. Upon closer inspection, our instance-wise entropy loss takes the following form, 

\begin{align*}
\begin{split}
    \mathcal{L}_{H, i } & = - \mathbf{\hat{y}}_i^T \ln (\mathbf{\hat{y}}_i) \\
    & = - \sum_{k = 1}^U \frac{\exp(\frac{1}{\tau} \mathbf{\mu}_k^T\mathbf{\Bar{e}_i)}}{\sum_{j =1}^U\exp(\frac{1}{\tau} \mathbf{\mu}_j^T\mathbf{\Bar{e}_i)}} \ln \frac{\exp(\frac{1}{\tau} \mathbf{\mu}_k^T\mathbf{\Bar{e}_i)}}{\sum_{j =1}^U\exp(\frac{1}{\tau} \mathbf{\mu}_j^T\mathbf{\Bar{e}_i)}}
\end{split}
\end{align*}

Which is minimized when the predicted class probabilities for the novel instance $ - \frac{\exp(\frac{1}{\tau} \mathbf{\mu}_k^T\mathbf{\Bar{e}_i)}}{\sum_{j =1}^U\exp(\frac{1}{\tau} \mathbf{\mu}_j^T\mathbf{\Bar{e}_i)}}; k = \{1,..., U\}$ follows a one-hot encoding vector upon satisfying that $\mathbf{e}$ aligns with the direction of one of the class representative hyperspherical prototype $\mu_k:k=\{1,..., U\}$.

The KL-divergence loss penalizes the empirical mean of the predicted probability. For entire novel data, the KL-divergence takes the following form,

\begin{align*}
\begin{split}
    \mathcal{L}_{kl-div} & = - \mathbf{p}_U^T \ln (\frac{1}{N^u}\sum_{i = 1}^{N^u}\mathbf{\hat{y}}_i) \\
    & = - \sum_{k = 1}^U p_{u, L+k}\ln (\frac{1}{N^u}\sum_{i = 1}^{N^u} \frac{\exp(\frac{1}{\tau} \mathbf{\mu}_k^T\mathbf{\Bar{e}_i)}}{\sum_{j =1}^U\exp(\frac{1}{\tau} \mathbf{\mu}_j^T\mathbf{\Bar{e}_i)}})
\end{split}
\end{align*}

The $\mathcal{L}_{kl-div}$ loss is minimized at the following condition

\begin{equation} \label{eq:lhs}
    \frac{1}{N^u}\sum_{i = 1}^{N^u} \frac{\exp(\frac{1}{\tau} \mathbf{\mu}_k^T\mathbf{\Bar{e}_i)}}{\sum_{j =1}^U\exp(\frac{1}{\tau} \mathbf{\mu}_j^T\mathbf{\Bar{e}_i)}} = p_{u,L+k} \mathtt{  } \forall k \in C_U
\end{equation}

We recall the Expectation Maximisation algorithm to fit the data in the movMF distribution where the mixing parameter converges in the following update equation \cite{gopal2014mises, banerjee2005clustering, rossi2022mixture} that is identical to the eq. \ref{eq:lhs}. 

\begin{equation} \label{eq:rhs}
   \pi_k =  \frac{1}{N^u}\sum_{i = 1}^{N^u} \frac{\exp(\frac{1}{\tau} \mathbf{\mu}_k^T\mathbf{\Bar{e}_i)}}{\sum_{j =1}^U\exp(\frac{1}{\tau} \mathbf{\mu}_j^T\mathbf{\Bar{e}_i)}}  \mathtt{  }  \forall k \in {1, ..., K}
\end{equation}

The key distinction from the EM-based movMF clustering approaches is that we have fixed the mean directions, and concentration parameter $\kappa = \frac{1}{\tau}$. Instead of learning the parameters, we train a deep learning function to project the classes to a pre-defined movMF distribution.

However, the order statistics and entropy constraints fail to capture the class notion. Instead, we rely on consistency loss for augmentation invariant embedding and supervised loss counterparts to align each class to separate mean direction consistently analogous to the uniform-alignment setting in self-supervised learning \cite{wang2020understanding}.

\section{Additional Experiments}
We report the performance of our additional experiments to demonstrate the robustness and potential limitations of our distribution learning-based NCD solution. The experiment settings overview is reported in Table \ref{table:exp_set}. Particularly, we perform experimentation from the labeled class group and dataset assumptions point of view. 

\begin{table}[!htb]
\caption{Additional experimentation settings}
\label{table:exp_set}
\centering
\footnotesize
\resizebox{\columnwidth}{!}{

\begin{tabular}{|l|l|l|l|l|}
\hline
Exp No & \multicolumn{1}{c|}{Dataset}                  & \multicolumn{1}{c|}{Labeled} & \multicolumn{1}{c|}{Novel}   & \multicolumn{1}{c|}{Comments}                                                                                                    \\ \hline
1      & \multicolumn{1}{c|}{\multirow{2}{*}{CIFAR10}} & \multicolumn{1}{c|}{Animal}  & \multicolumn{1}{c|}{Machine} & \multicolumn{1}{c|}{\multirow{2}{*}{Impacts of labeled classes}}                                                                 \\ \cline{1-1} \cline{3-4}
2      & \multicolumn{1}{c|}{}                         & Machine                      & Animal                       & \multicolumn{1}{c|}{}                                                                                                            \\ \hline
3      & SHAR                                          & 5 classes                    & 4 classes                    & \multirow{2}{*}{\begin{tabular}[c]{@{}l@{}}Over Cluster: Assuming more \\ number of classes than the true classes\end{tabular}}  \\ \cline{1-4}
4      & CIFAR10                                       & 5 classes                    & 5 classes                    &                                                                                                                                  \\ \hline
5      & SHAR                                          & 5 classes                    & 4 classes                    & \multirow{2}{*}{\begin{tabular}[c]{@{}l@{}}Under Cluster: Assuming less \\ number of classes than the true classes\end{tabular}} \\ \cline{1-4}
6      & CIFAR10                                       & 5 classes                    & 4 classes                    &                                                                                                                                  \\ \hline
7      & CIFAR10                                       & 5 classes                    & 5 classes                    & \multirow{2}{*}{\begin{tabular}[c]{@{}l@{}}Wrong Multinoulli \\ distribution assumption\end{tabular}}                            \\ \cline{1-4}
8      & SHAR                                          & 5 classes                    & 4 classes                    &                                                                                                                                  \\ \hline
\end{tabular}
}
\end{table}

\subsection{Impacts of labeled class notion:} 

\begin{figure}[!htb]
 \centering
 \includegraphics[width=\linewidth]{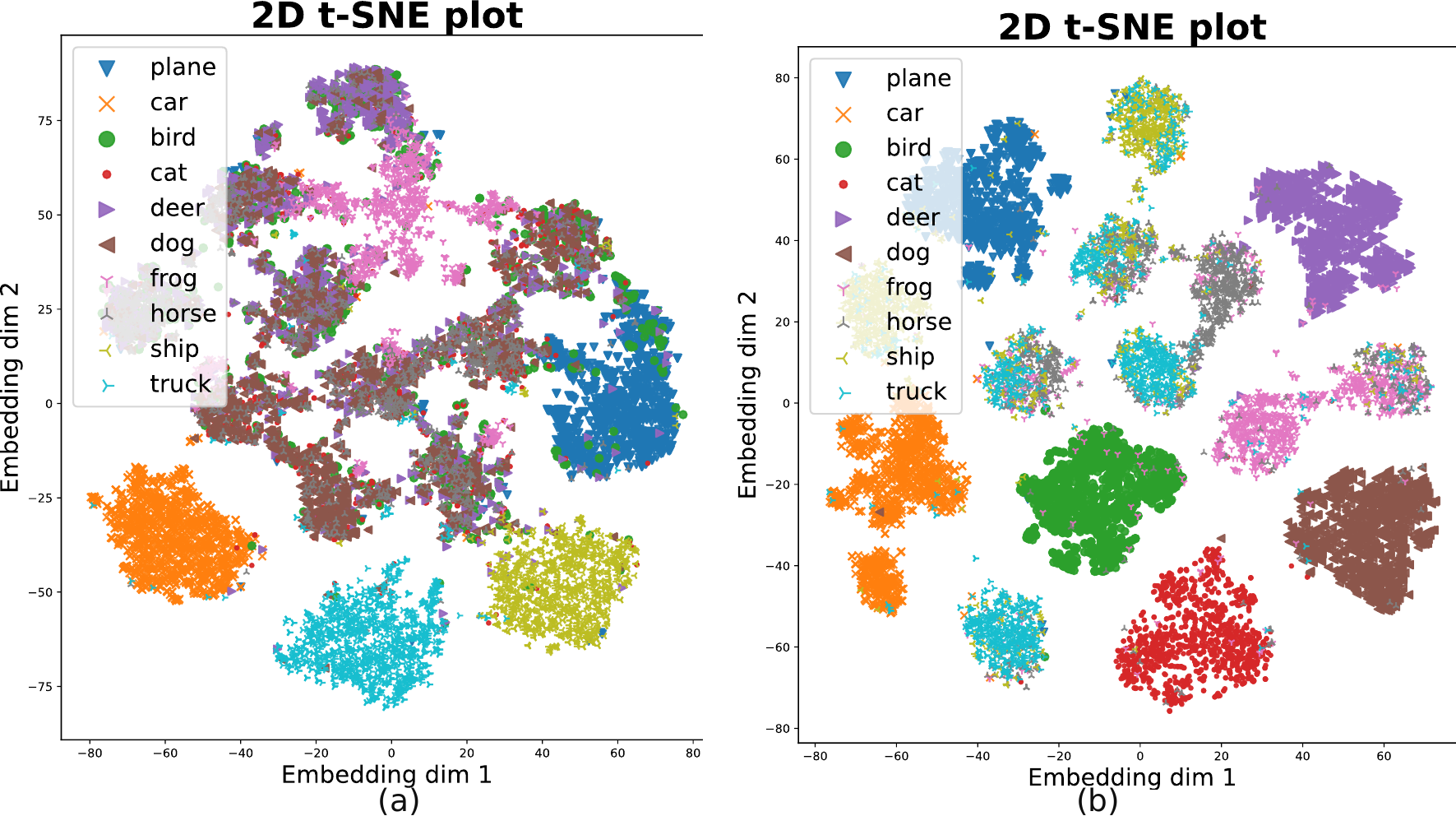}
 \caption{For the CIFAR10 dataset, two distinct instances of over-clustering are observed:
(a) In the Machine/Animal partitions, the model struggles to accurately recover the class structure.
(b) When considering the first five classes as labeled, the over-clustering performance becomes apparent.}
 \label{fig:oc_Cifar_comp}
\end{figure}

\begin{figure}[!htb]
 \centering
 \includegraphics[width=\columnwidth]{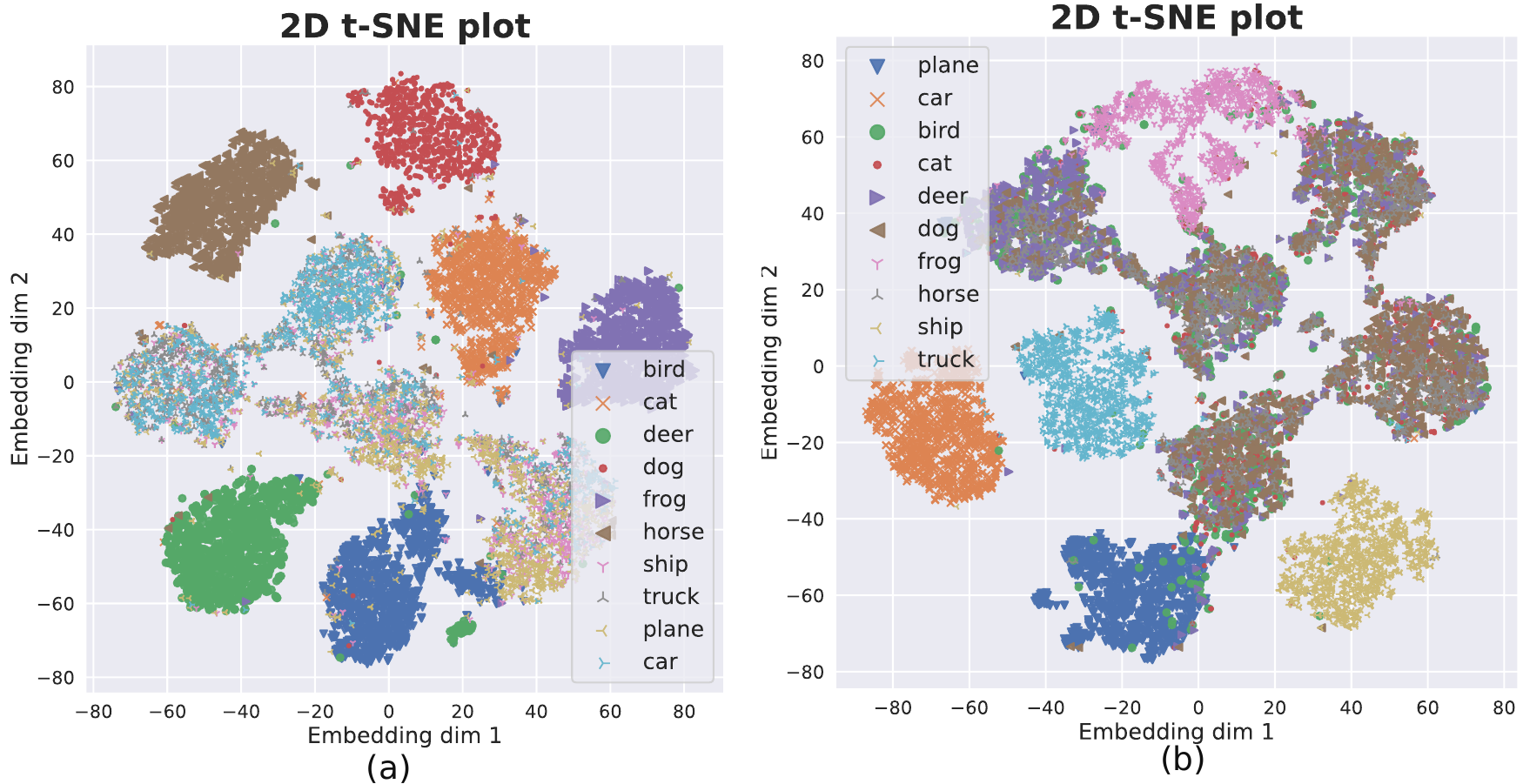}
 \caption{ResNet34 model demonstrates suboptimal performance when distinguishing between animals and machines:
(a) While the animal classes are labeled, machine classes are novel.
(b) Conversely, when the four machine classes are labeled, the animal classes are novel.}
 \label{fig:A_M_cifar}
\end{figure}

To delve into the influence of labeled class notions on our distribution learning-based NCD approach, we conduct two distinct experiments. In the first experiment, we designate the CIFAR10 Animal classes (bird, cat, deer, dog, horse, and frog) as the labeled set, while considering machine classes (ship, truck, plane, and car) as novel categories. Conversely, in the second experiment, we reverse the roles, with machine classes forming the labeled set and animal classes as novel categories. The model failed to separately represent based on the semantics of the novel classes in both settings as shown in Figure \ref{fig:A_M_cifar}. 
Furthermore, we observe that despite varying conditions, our model encounters challenges in achieving semantic-based clustering, even under scenarios of potential over-clustering, as visually demonstrated in Figure \ref{fig:oc_Cifar_comp}.

\begin{figure}[!htb]
 \centering
 \includegraphics[width=.6\linewidth]{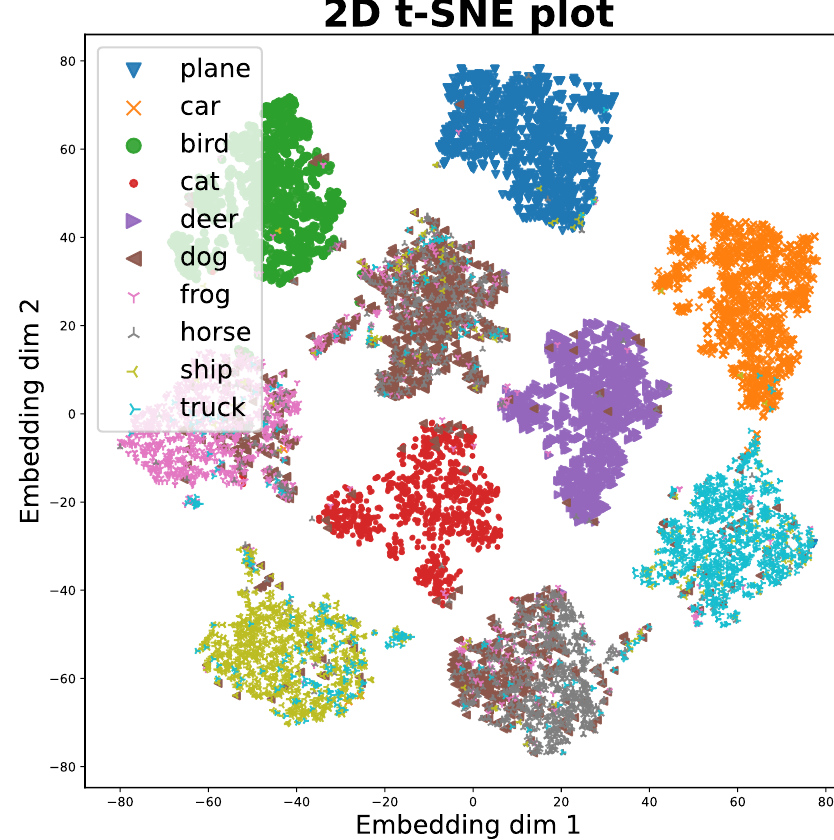}
 \caption{Suboptimal performance observed for the CIFAR10 dataset: Experimental data embedding reveals an overlap between the dog and horse classes.}
 \label{fig:tsne_plot_Mx_D_H}
\end{figure}

\subsection{Impacts of NCD data assumptions}

In this section, we delve into an analysis of the consequences arising from inaccurate assumptions concerning the novel class count and their corresponding distributions. We delineate three potential cases that emerge as a consequence of these erroneous data assumptions.

\begin{figure}[!htb]
 \centering
 \includegraphics[width=\columnwidth]{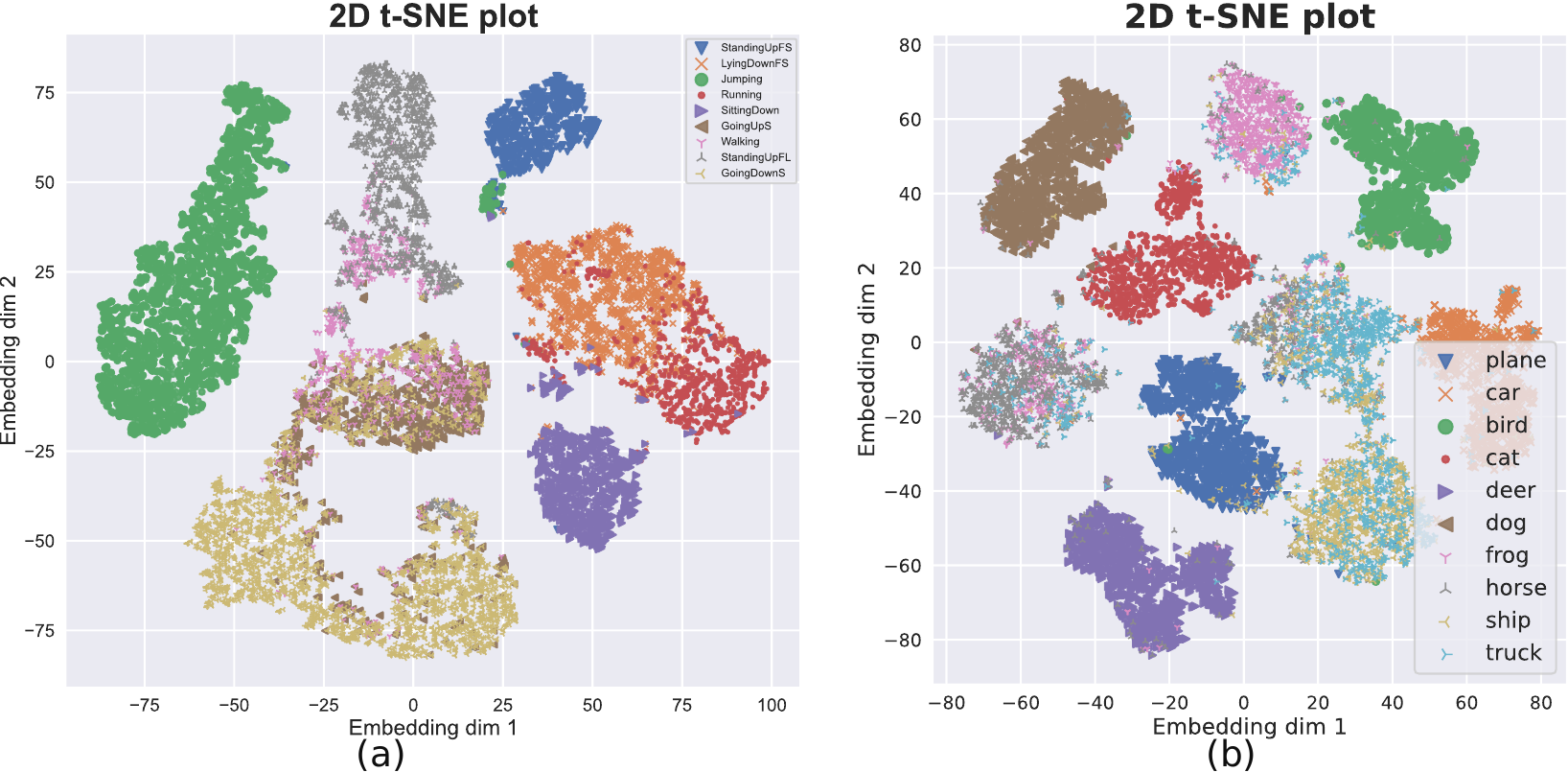}
 \caption{Representation overlap in the low-dimensional embedding space due to erroneous distribution assumption (a) SHAR (b) CIFAR10 dataset. The model underperforms in the tail classes.}
 \label{fig:wrong_dist}
\end{figure}

\textbf{Case 1: Wrong Distribution Assumption}: We consider uniform distribution assuming the unknown condition on the underlying novel class multinoulli distribution. The erroneous data distribution assumption enforces embedding representation overlap as the distribution loss optimization as visually demonstrated in Figure \ref{fig:wrong_dist} for two of the experimented datasets. 

\begin{figure}[!htb]
 \centering
 \includegraphics[width=\columnwidth]{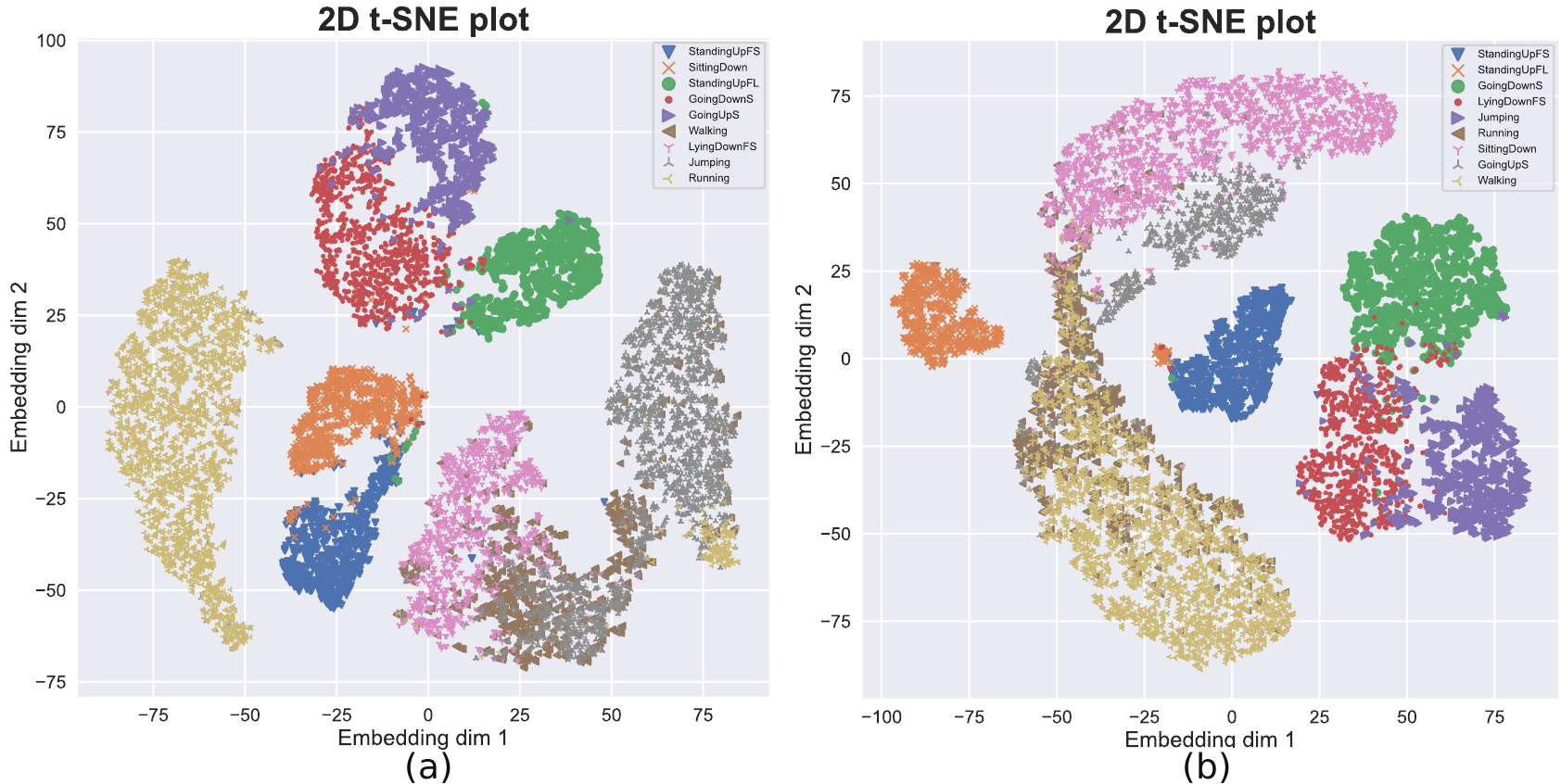}
 \caption{Performance of under clustering observed for the SHAR dataset:
(a) Assumption of two classes contrasts with the actual presence of four classes in one labeled/novel partition (walking and jumping overlap).
(b) A similar discrepancy is evident in another labeled/novel partition (walking and running overlap).}
 \label{fig:uc_SHAR}
\end{figure}

\begin{figure}[!htb]
 \centering
 \includegraphics[width=.6\columnwidth]{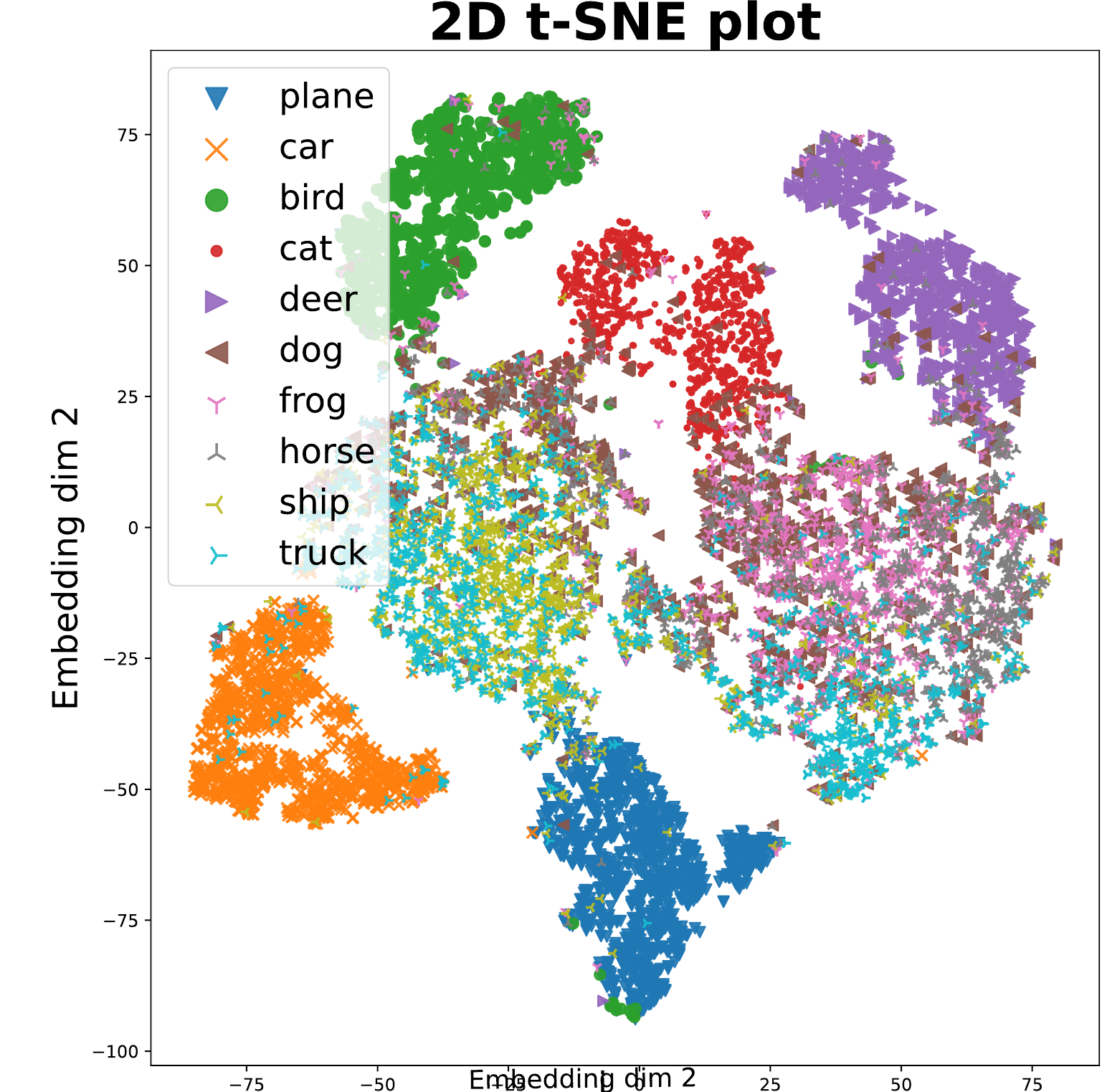}
 \caption{Suboptimal under clustering observed for the CIFAR10 dataset: While the true structure comprises five distinct classes, only two are assumed. Notably, the dog and horse classes collapse, forming a supercluster.}
 \label{fig:uc_cifar}
\end{figure}

\textbf{Case 2: Under-clustering Analysis}: 
We conduct our analysis under the assumption of a reduced count of novel classes relative to the total class count. Our experimental results using Cifar10 and SHAR datasets reveal that the clustering conditions lead to class collapse \cite{jing2021understanding}, compelling multiple classes to share the same embedding.
Interestingly, under balanced labeled/novel data partition, we observe semantic-driven class collapse, as visually depicted in Figure  \ref{fig:uc_SHAR} and \ref{fig:uc_cifar}. For example, we observe that similar semantic-based classes are collapsed together in the cifar10 dataset as depicted in Figure \ref{fig:tsne_plot_Mx_D_H}. 

\begin{figure}[!htb]
 \centering
 \includegraphics[width=\columnwidth]{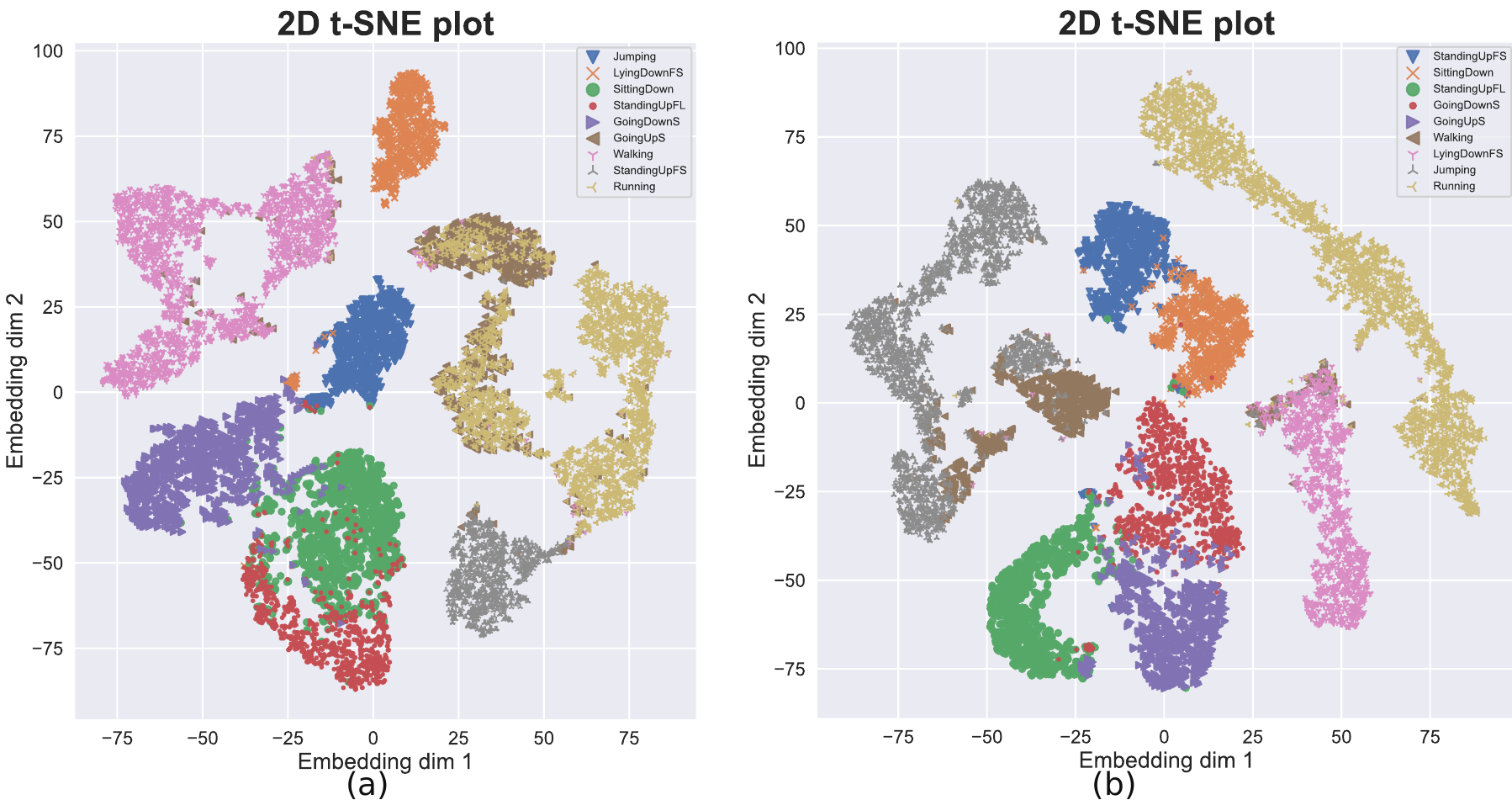}
 \caption{Over-clustering observed in two samples from the SHAR dataset across distinct partitions:
(a) The model adeptly identifies sub-clusters within the same action class. (b) A similar discrepancy is observed in another labeled/novel partition.}
 \label{fig:oc_SHAR}

\end{figure}

\begin{figure}[!htb]
 \centering
 \includegraphics[width=\columnwidth]{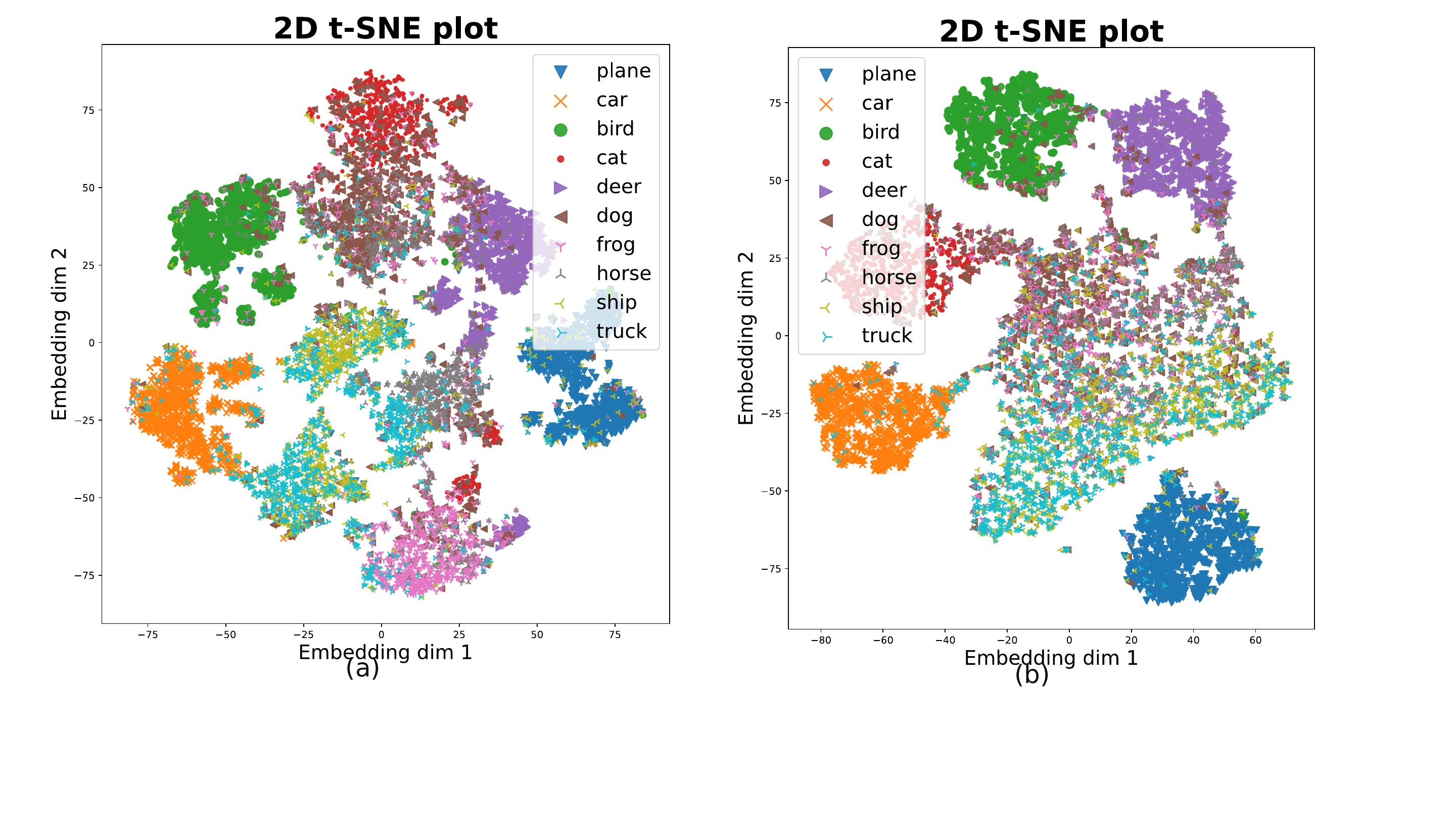}
 \caption{Over-clustering observed in two samples from the CIFAR10 dataset across distinct partitions:
(a) The model adeptly identifies sub-clusters within the same action class. (b) The model failed to converge to the specified number of classes.}
 \label{fig:oc_Cifar}

\end{figure}

\textbf{Case 3: Over-clustering Analysis} We perform over-cluster assuming a higher number of classes than the true class number.
Our investigation reveals that the model endeavors to disintegrate fine-grained class structures, emphasizing class-irrelevant features in an attempt to group data into a predetermined number of clusters as visually demonstrated in Figure \ref{fig:oc_SHAR} and \ref{fig:oc_Cifar}. 
However, it is noteworthy that the model encounters challenges when confronted with semantic shifts between labeled and novel classes. This deficiency becomes evident in Figure \ref{fig:oc_Cifar_comp} (a), where the model struggles to accurately identify novel classes due to the presence of semantic disparities.

\begin{figure}[!htb]
 \centering
 \includegraphics[width=\columnwidth]{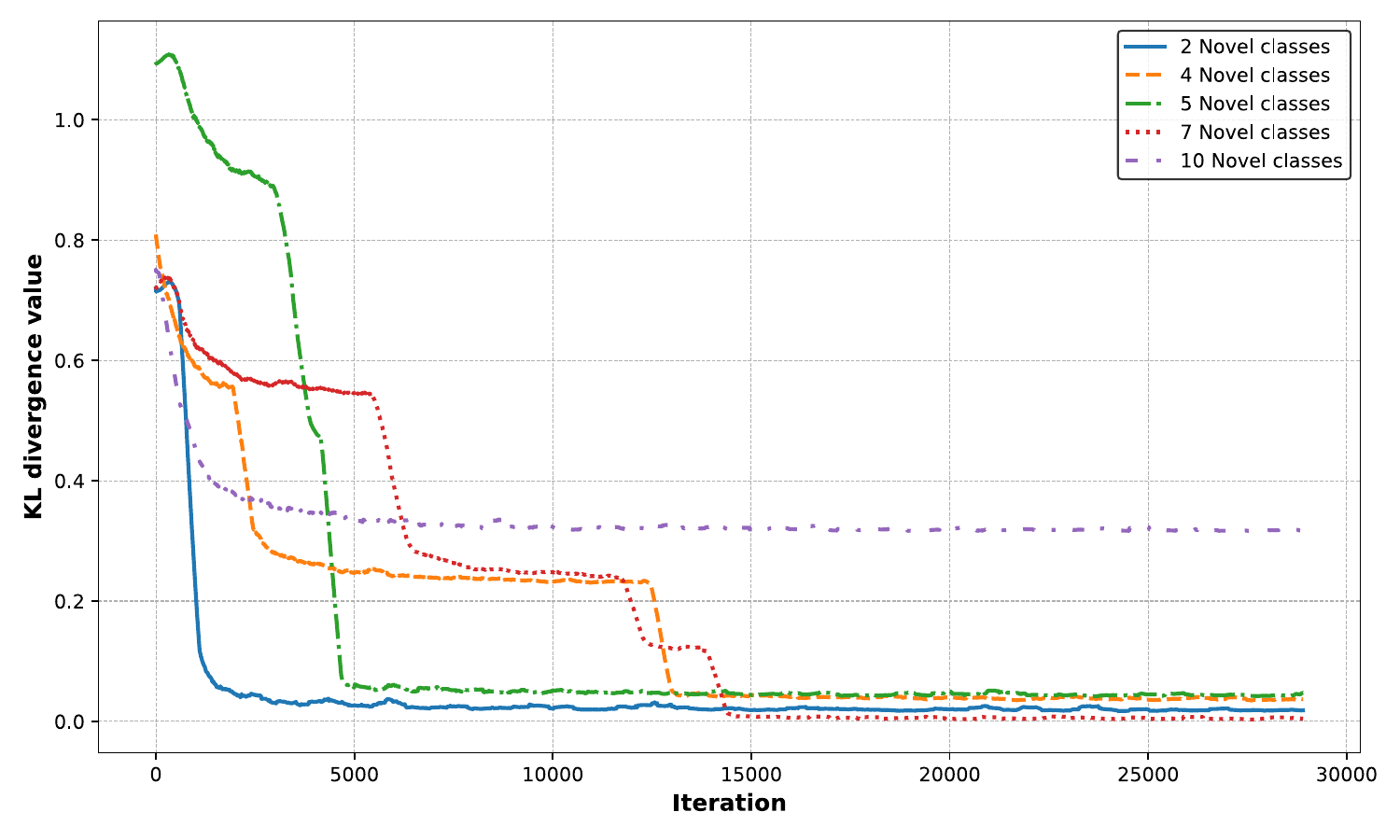}
 \caption{An analysis comparing the convergence of $\mathcal{L}_{kl-div}$  across varying numbers of assumed novel clusters. Notably, in every scenario, there are five genuine novel classes. CIFAR10 is used for this experiment.}
 \label{fig:kl_conver}
\end{figure}

We also observe the $\mathcal{L}_{kl-div}$ loss optimization properties for under-clustering and over-clustering and compare them with the baseline of right assumption settings. We report the average optimization curve in Figure \ref{fig:kl_conver}, obtained from three distinct runs for different weight initialization and subsequently smoothed for enhanced trend visualization.

\bibliography{refs}